\documentclass[twocolumn]{far}
\usepackage{far_techrpt}

\usepackage{microtype}
\usepackage{graphicx}
\usepackage{subcaption}
\usepackage{booktabs} 

\usepackage{hyperref}

\usepackage{amsmath,amsfonts,bm}

\def\eqref#1{equation~\ref{#1}}

\def\1{\bm{1}}

\DeclareMathAlphabet{\mathsfit}{\encodingdefault}{\sfdefault}{m}{sl}
\SetMathAlphabet{\mathsfit}{bold}{\encodingdefault}{\sfdefault}{bx}{n}

\DeclareMathOperator*{\argmin}{arg\,min}

\usepackage{cleveref}
\usepackage{url}
\usepackage{afterpage}
\usepackage{amsmath, amssymb, amsthm}

\newenvironment{figfullwidth}[2][hbtp]{%
\begin{figure*}[#1]%
\centering%
\def\@captiontext{#2}%
}{%
\caption{\@captiontext}%
\end{figure*}%
}

\newenvironment{tabfullwidth}[2][hbtp]{%
\begin{table*}[#1]%
\centering%
\caption{#2}%
}{%
\end{table*}%
}

\newenvironment{tabhalfwidth}[2][hbtp]{%
\begin{table}[#1]%
\centering%
\caption{#2}%
}{%
\end{table}%
}

\newcommand{\drcthree}{$\text{DRC}(3,3)$}
\newcommand{\drcone}{$\text{DRC}(1,1)$}
\newcommand{\resnet}{\text{ResNet}}

\title{\fontsize{28}{30}\selectfont Planning in a recurrent neural network that plays Sokoban}
\author{Mohammad Taufeeque,\quad Philip Quirke,\quad Maximilian Li$^*$,\quad\\Chris Cundy,\quad Aaron David Tucker,\quad Adam Gleave,\quad Adrià Garriga-Alonso \\
    \authoremail{\{taufeeque,adria\}@far.ai} \\
    \authorinstitution{FAR.AI, Berkeley, California, United States of America.\qquad $^*$Jane Street.}
}

\newcommand{\abstractwrapper}[1]{\abstract{#1}}

\newif\ificlrfinal
\iclrfinaltrue

\begin{document}
\maketitle
\logo

\newcommand{\aga}[1]{{\color{blue} TODO(aga): #1}}
\newcommand{\cc}[1]{{\small \color{green} CC: #1}}
\newcommand{\tf}[1]{{\small \color{red} TF: #1}}
\newcommand{\update}[1]{{\color{red} #1}}
\ifcsname definition\endcsname
\else%
\newtheorem{definition}{Definition}
\crefname{definition}{definition}{definitions}
\Crefname{definition}{Definition}{Definitions}
\crefformat{definition}{definition~#2#1#3}
\Crefformat{definition}{Definition~#2#1#3}
\crefrangeformat{definition}{definitions~#3#1#4--#5#2#6}
\Crefrangeformat{definition}{Definitions~#3#1#4--#5#2#6}
\fi%

\newtheorem{finding}{Finding}
\crefname{finding}{finding}{findings}
\Crefname{finding}{Finding}{Findings}
\crefformat{finding}{Finding~#2#1#3}
\Crefformat{finding}{Finding~#2#1#3}
\crefrangeformat{finding}{Findings~#3#1#4--#5#2#6}
\Crefrangeformat{finding}{Findings~#3#1#4--#5#2#6}

\crefname{section}{Section}{Sections}
\Crefname{section}{Section}{Sections}
\crefname{appendix}{Appendix}{Appendices}
\Crefname{appendix}{Appendix}{Appendices}

\abstractwrapper{
  Planning is essential for solving complex tasks, yet the internal mechanisms underlying planning in neural networks remain poorly understood. Building on prior work, we analyze a recurrent neural network (RNN) trained on Sokoban, a challenging puzzle requiring sequential, irreversible decisions.
  We find that the RNN has a causal plan representation which predicts 
  its future actions about 50 steps in advance. The quality and length of the represented plan increases over the first few steps. We uncover a surprising behavior: the RNN ``paces'' in cycles to give itself extra computation at the start of a level, and show that this behavior is incentivized by training.
  Leveraging these insights, we extend the trained RNN to significantly larger, out-of-distribution Sokoban puzzles, demonstrating robust representations beyond the training regime. We open-source our model and code, and believe
  the neural network's interesting behavior makes it an excellent model organism to deepen our understanding of learned
  planning.}

\section{Introduction}\label{sec:introduction}
In many tasks, the performance of both humans and some neural networks (NNs) improves with more
reasoning: whether by giving a human time to think before making a chess move, or by prompting or training a large
language model (LLM) to reason step by step \citep{kojima2022cot,chatgpto1}.

Among other reasoning capabilities, goal-oriented reasoning is particularly relevant to AI alignment. So-called
``mesa-optimizers'' -- AIs that have learned to pursue goals through internal reasoning \citep{hubinger2019risks} -- may
internalize goals different from the training objective, leading to goal misgeneralization
\citep{di2022goal,shah2022goal}. Understanding how NNs  learn to plan and
represent the objective could be key to detect, prevent or correct goal misgeneralization.

In this work, we focus on interpreting a Deep Repeating ConvLSTM \citep[DRC]{guez19model_free_planning} trained on Sokoban, a puzzle game often used as a planning benchmark \citep{peters2023solving}. We interpret the best network from \Citet{guez19model_free_planning}, \drcthree{}, with 3 recurrent layers that are applied 3 times per environment step. Further details of the network are provided in \cref{sec:setup}. We find that its internal plan representation \citep{tomsokoban} is causal, improves with more computation, and that the DRC learns to take advantage of that by often ``pacing'' to get enough time to refine its internal plan. We show similar results in \cref{app:results_on_drc11_and_resnet} for another DRC network and causal plan representation in a \resnet{} model.

\subsection{Definitions}
\Citet{guez19model_free_planning} showed that the DRC is very capable, generalizes well, and its performance improves with thinking steps at the beginning of an episode. Based on this, they claimed that DRC internally plans,
but did not make precise what this means. To clarify our contributions, we introduce a distinction between \textit{plans} and \textit{search algorithms}.

\begin{definition}[Plan]\label{def:plan-def}
A plan is a sequence of future actions $\{a_t\text{ for all }t>t_0\}$.
\end{definition}

\begin{definition}[Causally represented plan]\label{def:causality-of-plan}
Let $m\in\cal{M}$ be a simple model (e.g. logistic regression).
A plan is \emph{causally represented} if, for the current neural state $z_t$, and a hypothetical future environment state $s_k$, $k>t$:
\begin{enumerate}
\item \textbf{Prediction:}  we can extract the plan using $m$ with sufficient accuracy: $a_k \approx m(z_t, s_k)$ on average for $k > t$.
\item \textbf{Causality:} modifying $z_t$ alters the plan according to $m$, that is, preserves $a_k \approx m(z_t, s_k)$.
\end{enumerate}
\end{definition}

\begin{definition}[Search]\label{def:search}
A search algorithm is a decision-making process that involves generating multiple possible partial or complete plans, evaluating their predicted outcomes, and selecting the plan with the highest value.
\end{definition}

Unlike traditional planning algorithms, search in neural networks may use heuristics or partial models to evaluate plans, rather than specifying an exact world state they correspond to. The key feature of \emph{search} is the explicit representation and selection among competing plans. The competing plans do not have to be complete and could be a partial rollout.

\begin{definition}[Thinking steps]\label{def:thinking_steps} 
Timesteps where the agent receives the same observation repeatedly, with its predicted actions not executed in the environment. \end{definition}

\begin{definition}[Cycle]\label{def:cycle} A sequence of environment steps that starts and ends at the same state. \end{definition}

\subsection{Contribution statement}
\begin{finding}[The DRC causally represents its plan]
\label{hyp:plan}
The DRC maintains a plan that is consistently selected and causally represented in its hidden states. This plan predicts and drives the agent's behavior, and intervening on these representations alters the agent's actions. \emph{See \cref{sec:causal-plan}.}
\end{finding}
 Concurrently, \Citet{tomsokoban} also demonstrated \cref{hyp:plan} on handcrafted toy levels similar to the ones in \cref{app:toy-levels}, by intervening on a key step. We further test the causality of similar representations on random perturbations on difficult levels and propose a more precise causal intervention method, which works on any level and lets us alter any step in a plan.
 
\begin{finding}[Plan improves with computation]
\label{hyp:plan-improves}
With each internal iteration, the DRC network refines its plan, akin to doing a heuristic search, bringing it closer to the agent's eventual sequence of actions. \emph{See \cref{sec:plan-improves}.}
\end{finding}

\begin{finding}[Pacing behavior]
\label{hyp:pacing}
The DRC sometimes delays irreversible actions to allocate more computation time for refining its plan. This behavior is discovered during training and is not an artifact. \emph{See \cref{sec:agent_pacing}.}
\end{finding}

Evidence for~\ref{hyp:pacing}: the DRC's probed plans change 60\% faster during cycles compared to non-cycle steps. Additionally, training the DRC with a bonus for NOOPs (smaller than the per-step penalty) increases the proportion of NOOPs in cycle steps while keeping the total number of cycle steps the same. We believe \cref{hyp:pacing} is likely the correct explanation for \Citet{guez19model_free_planning}'s finding that forced thinking time improves performance.

\begin{finding}[Generalization to OOD larger levels.]
\label{hyp:larger_levels}
Using action probes, we are able to make the convolutional core of the DRC generalize to novel, significantly larger Sokoban puzzles beyond its training distribution. \emph{See \cref{sec:spatial-generalization}.}
\end{finding}
Previous work takes significant out-of-distribution generalization as evidence of algorithmic reasoning \citep{bansal2022endtoend,guez19model_free_planning}.

\begin{finding}[Lack of explicit search, low confidence]
\label{hyp:search} The \drcthree{} does not perform search by generating and evaluating multiple plans, before selecting the one with the highest predicted value.
\end{finding}
We were unsuccessful at finding a representation of the value of each plan using a value probe (\cref{app:probetraining}), and at correlating thinking steps with A* expanded nodes (\cref{fig:thinking-steps-vs-nodes}). It is possible that DRC performs search but with a different heuristic. The rest of the evidence is compatible with search, and with the DRC iteratively refining its plan with heuristics.

We demonstrate these findings on \drcthree{} in the main text. In \cref{app:results_on_drc11_and_resnet}, we show that the major Findings~\ref{hyp:pacing}$-$\ref{hyp:pacing} also hold true for \drcone{}, while only Finding~\ref{hyp:pacing} holds true for a \resnet{} model.




\textbf{Open-source resources.}
We open-source all models, tools, and interpretability data, offering a comprehensive resource for future research into planning behaviors\footnote{URL references removed during double-blind review. Code is available in the supplementary material.}. The DRC achieves an ideal balance of complexity and tractability for interpretability research with just 1.29M parameters.

{
  \begin{figfullwidth}[hbt]{
  \textbf{Top left:} Validation success rate on medium-difficulty levels solved vs.\ training steps, with varying numbers of initial thinking steps (forced NOOP actions) and a ResNet baseline. \textbf{Bottom left:}\label{fig:valid-curve-thinking-steps} Estimated planning effect (8-steps minus 0-steps) showing that planning emerges within 70M steps and increases for the hard levels (red) but decreases for medium levels (blue). \textbf{Middle:} Linear probe predictions for box movement direction. Green arrows indicate correct predictions and incorrect red arrows indicate plausible alternatives for Box 4. Opacity reflects frequency during the episode.
This probe causally affects agent actions as described in \cref{sec:causal-plan}. Boxes are numbered in their order of placement on targets.
 \textbf{Right:} The same level's input representation to the NN, where each pixel is a single tile. Walls are black, boxes are brown, targets are pink, and the robot is green.
  
  \label{fig:headline-figure}
  \label{fig:side-by-side-tinyworld}
}%
\includegraphics[width=0.4\linewidth]{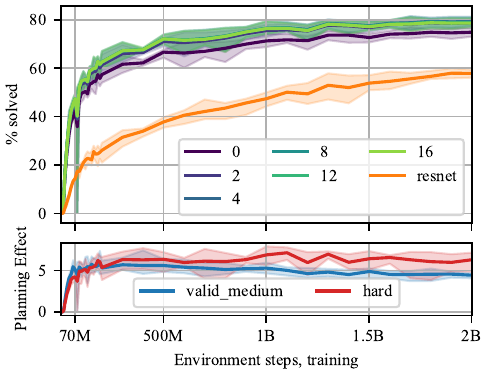}%
\hspace{0.01\linewidth}%
\raisebox{3.5mm}{\includegraphics[width=0.29\linewidth]{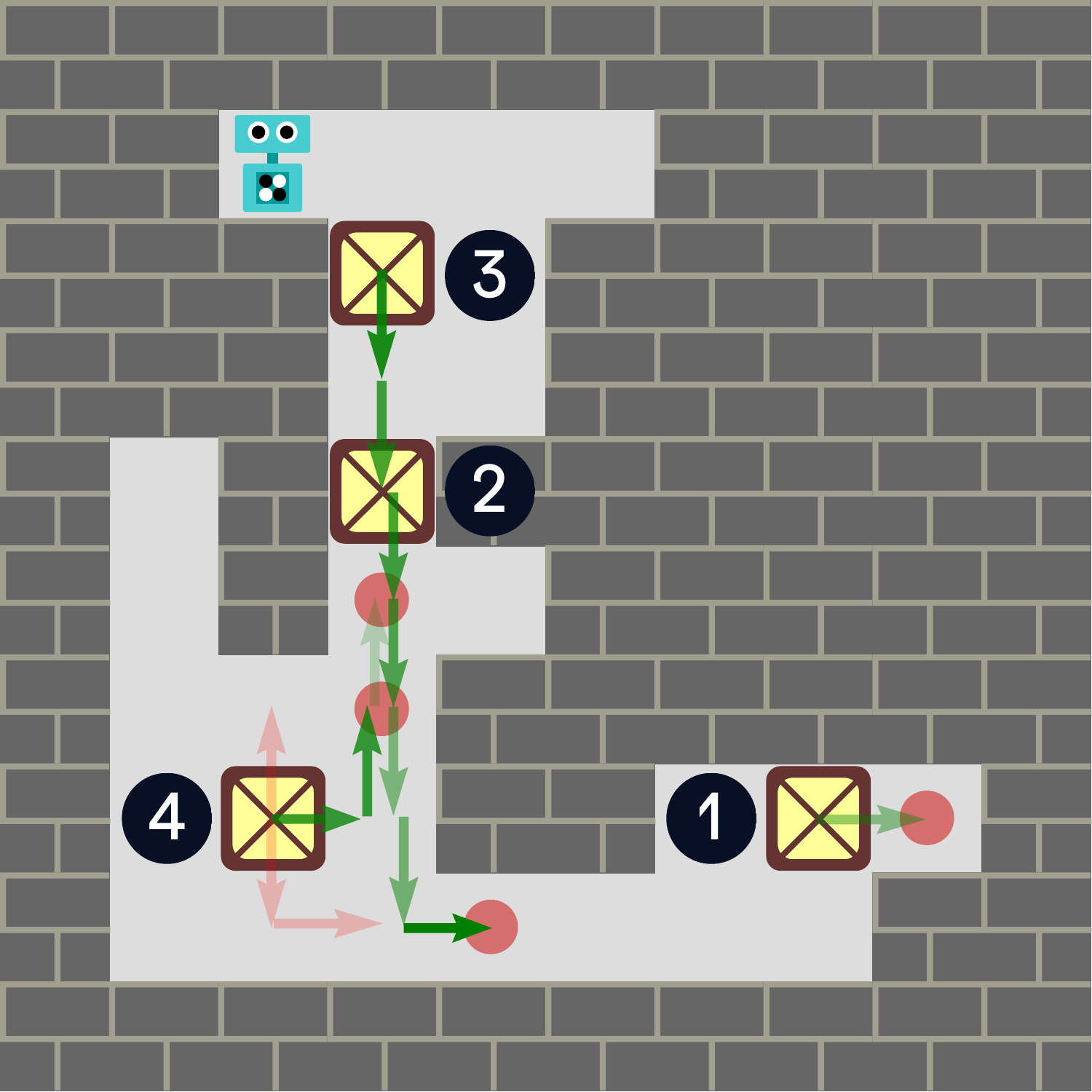}}%
\hspace{0.01\linewidth}%
\raisebox{3.5mm}{\includegraphics[width=0.29\linewidth]{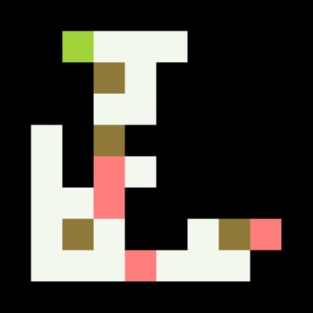}}%

\end{figfullwidth}
}

\begin{figfullwidth}[t]{%
\textbf{Left:} The \drcthree{} architecture  \Citep{guez19model_free_planning} with 3 convolutional layers repeated 3
times, embedded observation feeding into each layer, the last layer's output $h$ feeding back into the first layer, and
recurrent connections between layers. \emph{Blue:} Linear probes on hidden state grid cells to predict future actions
(\cref{sec:causal-plan}). \emph{Red:} Replacing the fixed-dimension MLP with probes enables the ConvLSTM core to
generalize to more challenging puzzles beyond the \(10\times10\) training grids
(\cref{sec:model-surgery-generalization}).
\textbf{Right:} $13 \times 17$ level from XSokoban-31 is solved by
    \drcthree{} after replacing the MLP.
    \label{fig:largest-level-solved-instantly}}
\hspace{0.01\linewidth}\includegraphics[width=0.50\linewidth]{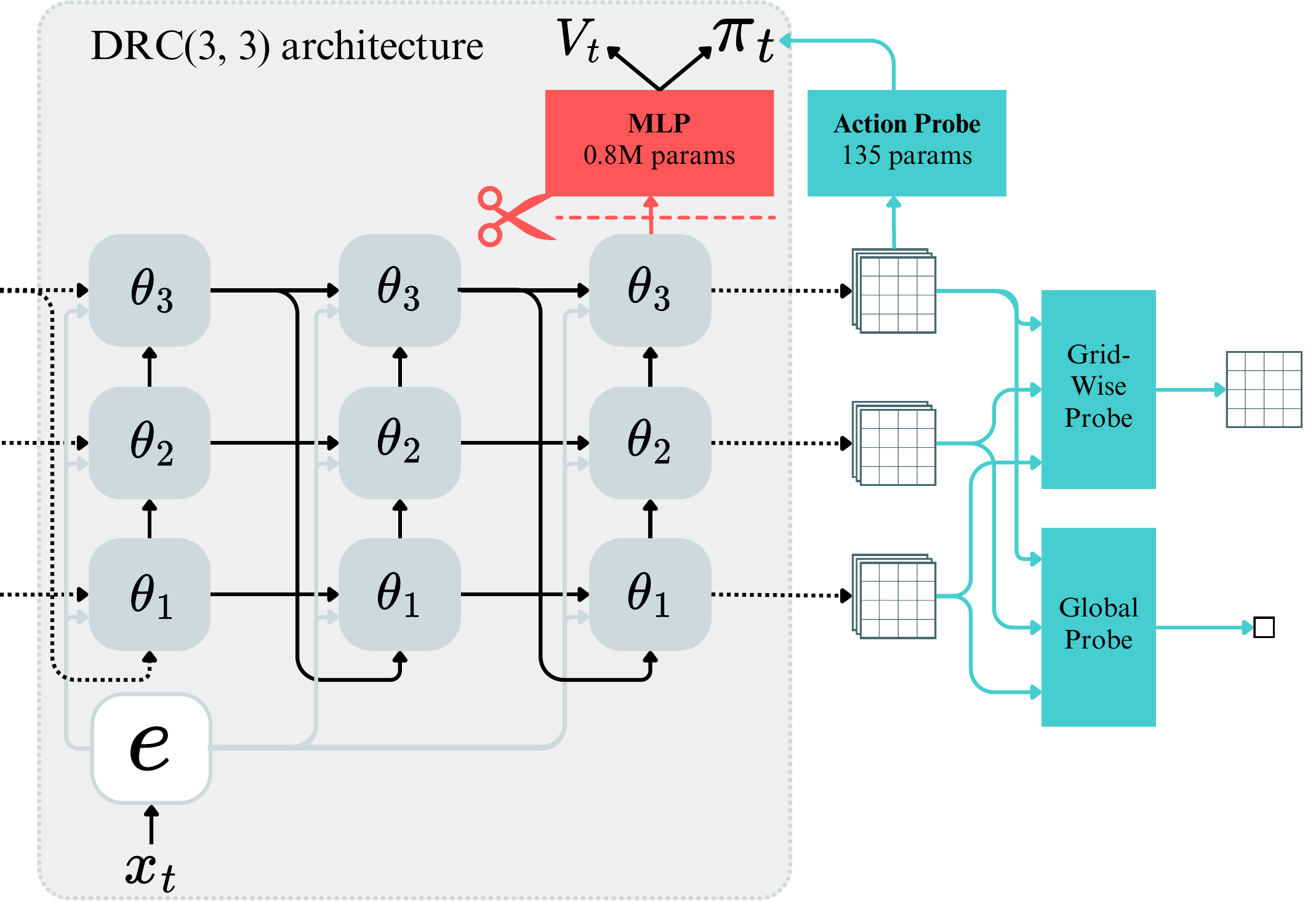}\hspace{0.03\linewidth}\includegraphics[width=0.445\linewidth]{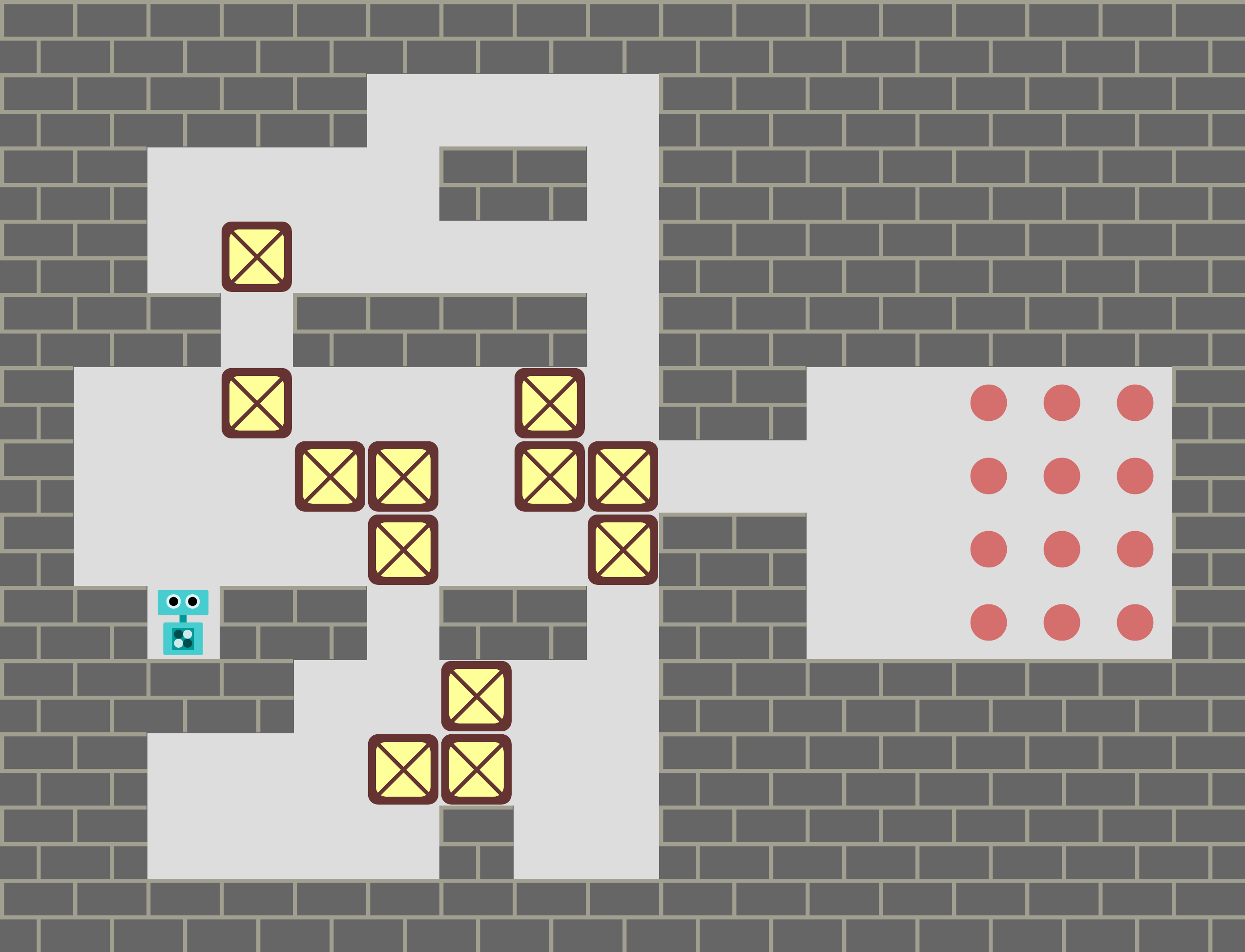}
\end{figfullwidth}

\section{Setting up the test subject}\label{sec:setup}
We train an agent closely following the setup from \Citet{guez19model_free_planning}, using the IMPALA V-trace actor-critic~\citep{espeholt18_impal} reinforcement learning (RL) algorithm with \citeauthor{guez19model_free_planning}'s Deep Repeating ConvLSTM (DRC)
recurrent architecture. We also train a ResNet baseline. For further architectural and training details, see \cref{app:training}.

\paragraph{$\text{DRC}(D, N)$ architecture.}
This paper primarily focuses on the behavior and representations of a \drcthree{} neural network
\citep{guez19model_free_planning}. The core component of this network is a $D$-layer ConvLSTM \citep{convlstm}, which is
repeatedly applied $N$ times per environment step (\cref{fig:largest-level-solved-instantly}, left). The output of the final ConvLSTM layer (the $D$th layer) is fed back into the input of the first layer at the next tick, effectively giving the network $D\cdot N$ layers of sequential computation to determine the next
action. In our setup, $D=N=3$ and $C=32$. \Cref{app:results_on_drc11_and_resnet} contains results on \drcone{} and a \resnet{} model.

A linear combination of the mean- and max-pooled ConvLSTM activations is injected into the next step, enabling quick
communication across the receptive field, known as (\textit{pool-and-inject}). An encoder block consisting of two
$4\times 4$ convolutions process the input, which is fed to each ConvLSTM layer. Each ConvLSTM layer's hidden states $(h, c)$ have the same number of channels $C$. Finally, an MLP with 256 hidden units
transforms the flattened ConvLSTM outputs into the policy (actor) and value function (critic) heads.

\paragraph{Dataset.} Sokoban is a grid-based puzzle game with walls, floors, movable boxes, and target tiles. The 
goal is to push all boxes onto target tiles while avoiding obstacles. We use the Boxoban dataset \citep{boxobanlevels},
consisting of $10 \times 10$ procedurally generated levels, each with 4 boxes and targets. The edge tiles are
always walls, so the playable area is $8 \times 8$. Boxoban separates levels into train, validation, and test sets with
three difficulty levels: unfiltered, medium, and hard. \citet{guez19model_free_planning} generated these sets by filtering
levels unsolvable by progressively better-trained DRC networks. So easier sets occasionally contain difficult
levels. In this paper, we train agents on the unfiltered-train (900k levels). For evaluation, we use the unfiltered-test
(1k levels)\footnote{We use unfiltered-test rather than unfiltered-validation to ensure direct comparability with
  \citet{guez19model_free_planning}.}, medium-validation (50k levels), and hard ($\sim$3.4k levels) sets, which do not
overlap. To test \drcthree{} generalization to different sizes, we use the levels collected by \citet{borgar-sokoban}
(see \cref{app:borgar-levels}).

\paragraph{Environment.} The observations are $10 \times 10$ RGB images, normalized by dividing each pixel component by 255.
Each tile type is represented by a unique pixel color \citep{SchraderSokoban2018}, illustrated in
\cref{fig:side-by-side-tinyworld} (right). The player has four actions available to move in cardinal directions (Up, Down, Left, Right). The reward is
-0.1 per step, +1 for placing a box on a target, -1 for removing it, and +10 for finishing the level by placing all of
the boxes. The time limit for evaluation is 120 steps, although large levels in \cref{sec:spatial-generalization} use
1000 steps. In the NOOP training part of \cref{sec:agent_pacing}, we modify the environment by adding an explicit NOOP action that has a fraction of the penalty as the other action.

\section{The \texorpdfstring{\drcthree{}}{DRC(3, 3)} causally represents its plan}
\label{sec:causal-plan}

We train probes\footnote{In the interpretability literature, probes are simple (usually linear) models that are trained to
predict specific labels (referred to as `concepts') from intermediate activations of a NN
\citep{alain16_under_inter_layer_using_linear_class_probes,belinkov2016probing}.  They provide a way to decode information represented in the network. However, that a probe can predict some information from NN activations is not enough to show that the NN is using that information. To show that, \citet{li2023emergent} argue that one has to intervene on the NN activations to change the probe output, \emph{and observe the NN's behavior changing accordingly.} } to predict the future actions of \drcthree{} and other features of the environment. Of these, the probe predicting the future move-direction of \emph{boxes} from every square in the grid has the strongest \emph{causal effect} on the actions of the DRC: intervening on the activations to change the probe's output also affects the future box directions \citep{li2023emergent}. Qualitatively, the sequence of box movements contains most of the information needed to solve Sokoban levels, so we consider this a \emph{plan}. Thus, we consider this strong evidence that the \drcthree{} represents and uses plans (\cref{hyp:plan}).
The spatial structure of the probes and some of the targets are adapted from \citet{tomsokoban}.
 
We find some evidence that the \drcthree{} considers multiple plans. For example, in the level in \cref{fig:headline-figure} (middle), the box probe initially predicts moving box 4 down and right (shown in red) but later revises this plan and takes different actions. Despite this, we have not found evidence of multiple \emph{simultaneous} plan representations or a mechanism to evaluate and decide between plans, such as a value comparison. Overall, this section provides strong evidence for a causally represented \emph{plan} (\cref{hyp:plan}), but very limited evidence supporting \emph{search} (\cref{hyp:search}).

\subsection{Probe methodology: interpretation and intervention}\label{sec:probes-and-steering}
\paragraph{Training.}
We train linear regression probes with L1 decay to predict environment features from the agent's activations. Similar to the concurrent work by \citet{tomsokoban}, we use two types of probes: \emph{(1) Grid-wise inputs} treat each square in the $10\times 10$ grid as a different data point, with inputs consisting of the $64$-dimensional LSTM state $(h, c)$ at a square, concatenated for the 3 layers. 
\emph{(2) Global inputs} aggregate information from the entire 10x10 grid, resulting in a 6400-dimensional input for each layer at every timestep.

The train and test dataset comprises states collected by evaluating the \drcthree{} on the hard Boxoban levels, excluding the first 5 steps of each episode as the plan is still forming. \Citet{tomsokoban} evaluated their probes on simple toy-levels similar to \cref{app:toy-levels}, which don't require long-term planning for most levels whereas we evaluate our probes on the hard set that can't be solved greedily and thus require long-term plans. For evaluation of multi-class probes, F1 scores are computed as one-vs-all: the presence or absence of a particular class is the probe label. We search the best learning rate and L1 decay with grid-search by evaluating the F1 on a validation split of $20\%$ of timesteps from the hard levels (\cref{tab:causal-probe-results}).

 \paragraph{Concepts (labels).} We choose concepts that are likely to encode steps in the agent's plan. Both the Agent-Directions and Box-Directions probes are grid-wise. These are central to our causality analysis and the same as \citet{tomsokoban}. We train five additional probes: Next-Box, Next-Target, Next-Action, Pacing, and Value, described in \cref{app:probetraining}.

 \begin{itemize}
\item \textbf{Agent-Directions probe.} Predicts the direction the agent takes from a square $(x, y)$ at the nearest future timestep. It has 5 outputs: NV (No Visit), UP, DOWN, LEFT, RIGHT. The probe takes the 192-dim hidden state activations concatenated across the 3 layers at a square $(x, y)$ as input. If the agent visits the square in the future, the probe predicts the corresponding direction; otherwise, it predicts NV. Thus, for each image observation, we get a $10 \times 10$ target.
\item \textbf{Boxes-Directions probe.} Same, but predicts the direction \emph{any} of the four boxes will move in at a given square.
\end{itemize}

\paragraph{Causal intervention.} To see if the NN uses these concepts in its algorithm, we edit the network's activations to influence probe outputs \citep{li2023emergent}, and measure the \emph{causal effect}: the fraction of cases where an intervention affects the agent's action accordingly.\footnote{This measurement is known in the literature as the \emph{average causal effect} or average treatment effect \citep{holland86}, the expected difference in a variable (whether the direction changes) when a causal intervention is present or absent.}. For a linear probe on activations \(h\) with parameter vector \(p\), we intervene to increase the logit \(p\cdot h\) to an adaptive value $\alpha$, following work on steering vectors \citep{turner2023activation, rimskySteeringLlamaContrastive2023, li2024inference}. More precisely, we find \(\argmin_{h'} \|h' - h\|_2^2\), with the constraint that \(h' \cdot p > \alpha\). The solution to this is \(h' = h + \hat{p}\max(0, \alpha - (h \cdot p))\), where $\hat{p}=p / \|p\|^2_2$. For multiclass direction probes, we also zero-out the logits of the other classes. For example, to influence the UP direction at a square, we set the UP logit to $\alpha$ and the DOWN, LEFT, RIGHT, and NV logits to 0. This intervention more precisely edits the network's representation as compared to the intervention of \(h' = h + p\) performed by \citet{tomsokoban} on toy-levels. Our intervention method is required to observe causal effects on standard difficult levels in the hard set.

{\small
\addtolength{\tabcolsep}{-0.1em}
\begin{tabhalfwidth}[t]{Causal and predictive probe results. Prediction column measures probe's F1 score to predict future information at all steps. Causal effect measures the fraction of cases where an intervention in the network's hidden state using probes affects the agent's action accordingly. Confidence is one of the mean estimator percentiles $[2.5\%, 97.5\%]$, whichever is furthest from the mean, estimated using 1000 bootstrap resamples. The \textsc{average} causal probe uses $24\text{k}$ data points for evaluation, and the \textsc{best-case} probe uses $8\text{k}$. 
\label{tab:predictive-probe-results}
\label{tab:causal-probe-results}}
\begin{tabular}{l|r|rrr}
\toprule
\textsc{Probe}
&\textsc{Pred (a)}
&\multicolumn{3}{|c}{\textsc{Causal Effect (b)}}
\\\midrule
& \textsc{F1}  & \textsc{$\alpha$} & \textsc{Avg} & \textsc{Best-case}\\
\midrule
  Box-Dir & $86.4 \pm 0.1$
 & 30 & $49.3 \pm 2.0$ & $82.5 \pm 2.5$ \\
  Agent-Dir & $72.3 \pm 0.1$
 & 10 & $7.1 \pm 0.3$ & $20.7 \pm 0.7$ \\
  Next box & $74.2 \pm 0.4$
 & 40 & $5.5 \pm 1.0$ & $15.1 \pm 2.5$ \\
  Next target & $54.3 \pm 0.5$
 & 30 & $4.6 \pm 0.8$ & $13.2 \pm 2.0$ \\
\bottomrule
\end{tabular}
\end{tabhalfwidth}
}


\subsection{Probe evaluation and causality}
\paragraph{Probe predictive power.} \Cref{tab:predictive-probe-results,tab:action_features} show that most probes demonstrate high predictive accuracy.

\Cref{fig:headline-figure} (right) and~\ref{fig:toy-levels-intervention} show visualizations of Agent-Directions and Box-Directions probes\footnote{The supplementary material provides visualization videos of all the probes across several
  levels.}. In particular, the box-directions probe can predict future box movements up to 50 steps in advance with over 90\% accuracy (\cref{fig:probe-steps-in-advance}, left).

\paragraph{Only the Box-Directions probe is strongly causal.} For each time step, we measure the causal impact of probes by intervening on the hidden state activations $h, c$ at the layer the probe was trained on. The results are reported in \cref{tab:causal-probe-results}(b), where we perform a grid-search over the intervention strength $\alpha$ to find the optimal value. Despite our adaptive scaling, high values of $\alpha$ can still destabilize the agent's behavior, causing random actions, while low values of $\alpha$ may fail to induce the intended behavior. Most probes show little
causal effect, the Box-Directions probe (used in \cref{fig:headline-figure}) demonstrates strong causality, and the Agent-Directions probe shows mild causality.

Even then, the Box-Directions probe seems to alter box movements only when they \emph{do not lead to naive deadlocks}. For instance, the model avoids pushing a box into a wall if later steps indicate that it would need to get the box off the wall to get it to the target. To account for this behavior, we introduce the \emph{Best-case causal effect} in
\cref{tab:causal-probe-results}(b). This tests all three alternate directions that a box could move, counting the probe as ``causal'' if it successfully influences \emph{any} of these directions without causing a naive deadlock.

By intervening on the probes, we can lock the DRC into a plan when there are two equally valued options. \Cref{fig:toy-levels-intervention} in \cref{app:toy-levels} shows visualizations of how causal interventions with the directions probe alter the trajectory of the agent.
However, in most cases, if we stop intervening with a suboptimal plan, the DRC formulates a better plan online and follows it. This and the naive deadlock problems may be caused by us intervening only on the current square of the box, target or agent: the NN activations still contain contradictory information about other squares, so the NN malfunctions.

\section{The plan improves with computation}\label{sec:plan-improves}
\Citet{guez19model_free_planning} showed that the DRC solves more levels if we give it thinking steps at the start of episodes (see \cref{def:thinking_steps}). But what happens in the network during these thinking steps? Using the Box-Directions probe from \cref{sec:causal-plan}, we show that the represented plan increases in length and accuracy (\cref{fig:thinking-steps-success-plan-qual}, middle and right), demonstrating \cref{hyp:plan-improves}. Extra thinking is more helpful for levels that have longer optimal solutions (\cref{fig:thinking-steps-vs-optimal}, right), or which require less myopic thinking to place the first few boxes (\cref{fig:steps-to-box}, left)(b). Thus, we are moderately confident that extra thinking improves the success rate by letting the network develop its plan long enough that it finds a non-myopic solution instead of accidentally locking the level.




\begin{figfullwidth}[t]{\textbf{Left:} Average time step to place each box $B_i$ on target for different numbers of thinking steps. \textbf{(a)} Averages across all levels. \textbf{(b)} Averages over levels solved by $6$ thinking steps but not solved by $0$ thinking steps. More thinking steps make the DRC avoid greedy strategies in favor of long-term return. The 95\% confidence intervals computed with the bootstrap method over the levels is very small and not visible in the plot. \textbf{Right:} Average optimal solution length of levels grouped by the number of thinking steps at which the level is first solved. Levels that take longer to solve tend to be harder. NS stands for ``not solved''.%
\label{fig:steps-to-box}%
\label{fig:thinking-steps-vs-optimal}%
}
\includegraphics[width=\linewidth]{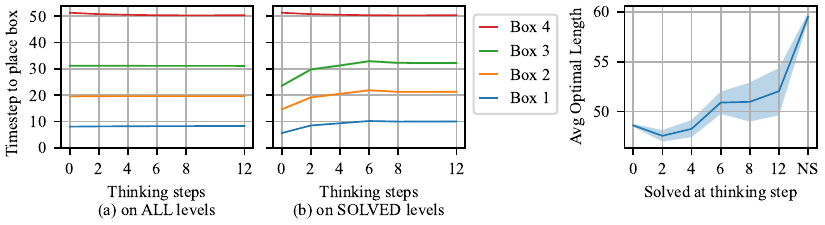}
\end{figfullwidth}

\paragraph{Effects of thinking time: non-myopic network}
We first examine how additional thinking time impacts the DRC performance by performing thinking steps at the start of an episode, allowing the network to
process its hidden state. As shown by \citet{guez19model_free_planning}, adding 6 thinking steps increases the success rate by $4.7\%$, with a slight drop beyond 16 steps
(\cref{fig:success_rate_with_thinking}).

This improvement is more pronounced for harder levels, which require longer optimal solutions (\cref{fig:thinking-steps-vs-optimal}, right). Notably, thinking time does not correlate with the number of nodes expanded by an A* search, suggesting the DRC works differently (\cref{fig:thinking-steps-vs-nodes}).

Thinking reduces early error in levels where placing the first box may block completion. \Cref{fig:steps-to-box} (left) (b) shows that without extra thinking steps, the agent pushes the first box to a target about 4 steps too early, and notably much earlier than average (\cref{fig:steps-to-box}, left). Providing 6 thinking steps prevents this mistake. While some improvement stems from avoiding catastrophic moves, the fact that the average time to push the first box exceeds 6 steps even without forced thinking suggests that the network uses the additional time to form a more complete solution, which changes behavior.

The impact of thinking time varies with training duration and level difficulty. \cref{fig:headline-figure} (bottom-left) shows that most of the planning improvement occurs within the first 70M steps. Beyond this point, the planning effect grows a bit more until 200M steps and then stabilizes (hard levels) or slowly decays (medium validation). Based on this, we speculate that the network develops better heuristics for when to think on medium-difficulty puzzles, reducing its default myopia.


\paragraph{Plan improvement.}
If DRC refines its plan with extra computation (\cref{hyp:plan-improves}), probe-extracted plans should become more predictive of the agent's actions as more computation time is provided. \Cref{fig:chain} (right) confirms this effect: the plan initially has low accuracy and rapidly improves as DRC is given more thinking steps. Additionally, \cref{fig:chain} (middle) shows that the plans increase in complexity over time, with the length of the plans, defined as sum of length of continuous chains starting from boxes, growing significantly during these additional computation steps.

\begin{figfullwidth}[t]{\textbf{Left:}\label{fig:probe-steps-in-advance} The number of steps in advance that the Box-Directions probe is able to predict a box move. The probe predicts $\ge 90\%$ of the actions 50 steps before they occur. \textbf{Middle and Right:} The plan quality, as measured by summing length of probe-predicted chain of box-directions starting from boxes and counting squares with positive non-empty predictions, and the F1-score increases over thinking steps (\cref{sec:probes-and-steering}). This suggests that the DRC refines its plan during computation, including across the three ticks per environment step.\label{fig:thinking-steps-success-plan-qual}\label{fig:chain}}
\includegraphics[width=\linewidth]{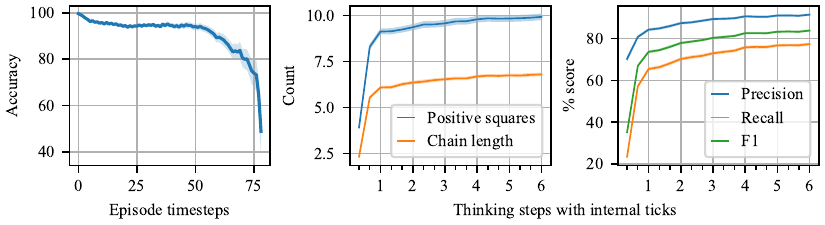}
\end{figfullwidth}


\section{Purposeful ``pacing'' to get more computation}\label{sec:agent_pacing}
On occasion, the DRC exhibits a curious behavior: the agent ``paces'' in a cycle, returning to the same location, without touching any box. This behavior does not advance the puzzle state, so it is  sub-optimal due to the per-step penalty incurred. Why is it still present in the trained network?

We hypothesize (\ref{hyp:pacing}) that this behavior is incentivized by training, and its purpose is to give the DRC enough time to develop the plan. We show this with multiple lines of evidence, based on observing and intervening on NN behavior, and speed of change in the plan during cycles. We also re-train the \drcthree{} with smaller penalties for taking NOOP actions, and show that the total number of cycles stays constant while the number of NOOPs increases. Through all these experiments, we take care to highlight that most cycles are deliberate for plan computation, as opposed to a small fraction of cycles that could be occurring accidentally

Unless otherwise noted, the following experiments are for cycles recorded on medium-validation levels. We merge overlapping cycles, resulting in a total of $13\,702$ cycles.

\paragraph{Cycles happen early in levels.} In some levels, a single sub-optimal step can render the puzzle unsolvable, so it is very important to develop enough of the plan to prevent that. \Cref{sec:plan-improves} showed that extra thinking steps help in large part by preventing these mishaps, so it makes sense that training would incentivize doing similarly early in a level. Accordingly,
\cref{fig:cycle-hist-and-replace-with-thinking} (left) shows that most cycles start in the first 5 steps of the episode.

\paragraph{Plans improve much faster in cycles.}  If the DRC uses cycles to refine its plan, we expect the plans to be worse at the start of cycles (because they need improvement), and to change faster during the cycle. We check this by looking at the F1 score of box-directions probe at the first step of cycles, and the rate of change in F1 and plan length found with box-directions probe, as measured by number of positive predictions of the probe.

Plan length and quality are highly influenced by whether the agent is close to the beginning or end of a level. To account for this, we pair each step in a cycle with a non-cycle step from a random episode, such that their timesteps $t$ match.

Using the hard-level set, we find that the F1 score at cycle start points is $51.13\% \pm 1.52\%$, compared to $68.13\% \pm 1.45\%$ for non-cycles. Per step, the F1 score improves on average by $0.96\% \pm 0.15\%$ during cycles, versus $0.60\% \pm 0.15\%$ during non-cycles. Finally, the plan length (number of squares with a predicted action) grows by 2.03 squares per step during cycles, compared to 1.37 squares per step during non-cycles.

\Cref{fig:cycle-illustration} shows the distribution corresponding to some of these aggregate statistics. While cycle-step improvements definitely skew larger, the distributions has some overlap, which suggests that some cycles could be accidental in which F1 improvement is not larger than non-cycles steps.

\paragraph{Training with smaller NOOP penalty.} The original version of the \drcthree{} could not take NOOPs, so it has to pace to get time to think. We train for 800M steps a version of the \drcthree{} which gets a smaller penalty for taking a NOOP than for moving. If the DRC is deliberately stalling when the plan is not good enough, we would expect it to use NOOPs when it is stalling \emph{deliberately}. At the same time, the total number of cycles should stay constant: each step incurs a penalty, so it is still optimal to solve the level as quickly as possible.

\begin{figure}[!t]
    \centering
    \includegraphics[width=\linewidth]{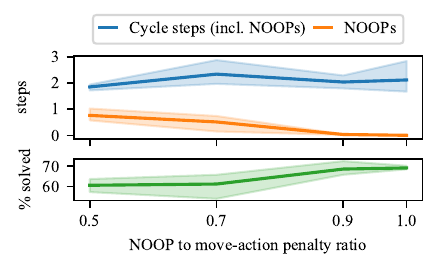}
    \caption{\textit{Top:} Average number of NOOPs and cycle steps per episode, when training with a NOOP action that has less penalty than the movement actions. This makes the network substitute many of the cyclic steps with NOOP steps on average, indicating it is a deliberate behavior. The total number of cycle steps (blue) stays constant. \textit{Bottom:} percentage of validation-medium levels solved for each network. Error bars show the minimum and the maximum over 3 training seeds.}
    \label{fig:noop-reward}
 \end{figure}

\Cref{fig:noop-reward} shows exactly this effect. The per-move penalty is fixed to $0.1$, and the NOOP penalty varies in $\{0.05, 0.07, 0.09, 0.1\}$. We plot the total number per episode of steps in cycles (including NOOPs) and of NOOPs. As expected, the NOOPs per episode increase with smaller NOOP penalties, and the cycle-steps per episode stay roughly constant (slightly decrease). At the same time, the $\drcthree{}$ performance stays similar: it goes from solving $70\%$ to $60\%$ of levels. The trend of more cycles getting replaced with NOOPs suggest that most cycles are deliberate, but a small fraction could be accidental that occur anyways. We exclude cycles that happen after the last box is placed in a level because when the DRC sees a level is locked, it goes in cycles until the time limit, and that would make the results above largely depend on the time limit.

\paragraph{Cycles can be substituted by NOOPs.} \Cref{fig:cycle-hist-and-replace-with-thinking} (middle) shows that forcing the model to think for six steps eliminates about 75\% of the early cycles. Thus, we can again conclude that most cycles are deliberately performed by the network. We can also directly replace would-be cycles with NOOPs. Just when the DRC would start a cycle of length $N$, we intervene and make it take $N$ thinking steps. \Cref{fig:cycle-hist-and-replace-with-thinking} (right) shows that these thinking steps largely replace the cycles: in at least $60\%$ of the levels, the DRC followed the same trajectory for a minimum of $30$ steps. For comparison, the median solution length for train-unfiltered is exactly 30 steps.




\begin{figfullwidth}[t]{\textbf{Left:} Histogram of cycle start times on the medium-difficulty validation levels. \textbf{Middle:} Total cycles the agent takes in the first 5 steps across all episodes with $N$ initial thinking steps. \textbf{Right:} Proportion of trajectories that match when replacing $N$-length cycles with $N$ thinking steps, tracked for $x$ steps post-cycle.\label{fig:cycle-hist-and-replace-with-thinking}}
\includegraphics[width=\linewidth]{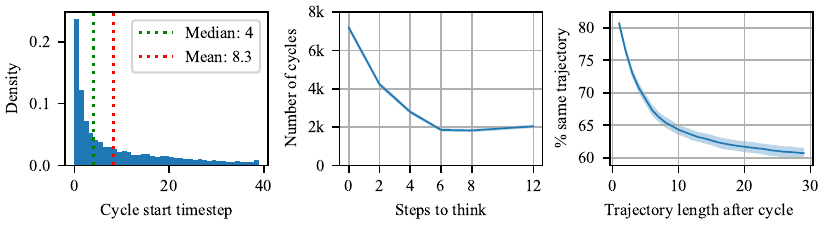}
\end{figfullwidth}

\section{Generalizing beyond \texorpdfstring{$10 \times 10$}{10 x 10} inputs and training examples}
\label{sec:model-surgery-generalization}
\label{sec:spatial-generalization}

We extend the DRC's capabilities by modifying its internal structure to generalize beyond the original $10 \times 10$ input size and four-box training distribution. Analysis of the ConvLSTM core reveals that layer 3, at the final tick, encodes four key channels representing the next action provided in \cref{tab:action_features}. Since the ConvLSTM is completely convolutional, it can process inputs of \emph{arbitrary} sizes, unlike the full DRC architecture, which flattens the convolution output and passes through an MLP layer.

\paragraph{Spatial aggregation.} Using layer 3 next-action probes (\cref{tab:action_features}), we predict at \emph{each} spatial location the next action of the DRC. To turn the probe's grid-wise outputs into a single action prediction, we need to aggregate them spatially. We linearly combine three spatial aggregation methods: mean-pooling, max-pooling, and the proportion of positive probe outputs. The optimal combination weights and a bias for each action are learned by minimizing cross-entropy loss against the MLP block using the Adam optimizer on 3000 levels (1000 from each Boxoban training set).

\paragraph{Results.} The aggregated features achieve $83\%$ accuracy on the training set and $77.9\%$ on medium-difficulty validation levels. Although lower than individual probes F1 scores \cref{tab:action_features}, these enable the adjusted ConvLSTM to solve many out-of-distribution levels, including larger grids and more than four boxes, e.g., \cref{fig:largest-level-solved-instantly} (right) and \cref{fig:successful-larger-levels}. Specifically, from the test set by \citet{borgar-sokoban}, it solves $64/484$ levels with both dimensions $>10$, $38/203$ levels with only one dimension $>10$, and $221/486$ levels within the original constraints. A detailed breakdown by level collection of performance and largest level solved is provided in \Cref{app:borgar-levels}. The largest level solved per level-collection is between 3-4x larger in area and number of boxes compared to the $10 \times 10$ levels with 4 boxes seen during training.

Our findings show that by aggregating interpretable spatial features, the ConvLSTM generalizes to larger and more complex puzzles without retraining. This suggests that much of the DRC's planning ability is encoded in the ConvLSTM core, and refining spatial predictions allows for scalable generalization in neural planning systems.

\paragraph{Stress-testing on zig-zag.} How well does this extension to larger levels work? We stress the network by placing it in straightforward (but long) $n \times n$ zig-zag levels (\cref{fig:zig-zag-level}). The \drcthree{} solves all levels for $n \le 15$, but fails at $n \ge 16$ no matter how many thinking steps.

\section{Related Work}

Our work contributes to the growing field of mechanistic interpretability focused on understanding reasoning and planning processes within neural networks. While the internal mechanisms of complex agents remain largely opaque, several lines of research provide context for our investigation.

\paragraph{Interpreting planning and reasoning in neural networks.}
Prior studies have investigated planning mechanisms in simpler settings like mazes \citep{mini23_under_contr_maze_solvin_polic_networ, knutson24_logic_extrap_mazes_with_recur_implic_networ, brinkmann24mechanal}, gridworlds \citep{decision-transformer-interpretability}, and graph search problems \citep{ivanitskiy2023linearly}. Others have explored reasoning in Large Language Models (LLMs) on tasks such as block-stacking \citep{men24_unloc_futur}. These works often focus on identifying representations of state or specific concepts, rather than studying the representation involved in planning. Researchers have also investigated sophisticated game-playing agents like Leela Chess Zero \citep{jenner24_eviden_learn_look_ahead_chess} and AlphaZero \citep{McGrath2021AcquisitionOC, schut23_bridg_human_ai_knowl_gap}, uncovering evidence for lookahead mechanisms or high-level local concepts important for the game. Our work builds on these efforts but specifically targets the explicit representation and causal role of \emph{plans} (complete sequences of future actions, \cref{def:plan-def}) within the DRC network trained on Sokoban, a game demanding complex, foresightful planning.

\paragraph{Sokoban planning and the DRC agent.}
The DRC network trained on Sokoban provides a compelling case study of long-term deliberate planning owing to its ability to utilize extra thinking time \citep{guez19model_free_planning}. Our work aims to understand and explain this behavior by interpreting the DRC's internal states. Concurrently with our research, \citet{tomsokoban} also interpreted a DRC agent trained on Sokoban, by introducing probes similar to ours that predict future box and agent movements and found evidence for plan representations (similar to our \cref{hyp:plan}). Our work complements and extends theirs in several key ways. \Citet{tomsokoban} performed causal interventions on $200$ handcrafted toy levels, whereas we perform more rigorous causal tests on $24k$ datapoints sampled from difficult standard benchmark levels and devise an intervention methodology that edits the representations more precisely (\cref{sec:causal-plan}). Additionally, our work yields novel findings: \cref{hyp:larger_levels} offers further evidence from probe-guided OOD generalization that DRC's planning capability generalizes to larger levels not supported by the architecture, while \cref{hyp:pacing} on pacing behavior provides explanation for the emergence of the performance improvements through ``thinking time''.


\paragraph{Probing and intervention methods.}
Our methodology relies on probing internal activations to decode information embedded linearly and intervening on these activations to test causality. Linear probing has been widely successful in finding representations of game states \citep{li2023emergent, nanda23_emerg_linear_repres_world_model, karvonen24_emerg_world_model_laten_variab} or spatial information \citep{wijmans2023emergence} in various networks trained on game sequences.

\paragraph{Adaptive computation for reasoning.}
The finding that the DRC's plan improves with computation (Finding~\ref{hyp:plan-improves}) and that the agent learns to "pace" (Finding~\ref{hyp:pacing}) connects to research on improving neural network reasoning by altering training setup or architecture and leveraging adaptive computation time in RNNs \citep{schwarzschild21_can_you_learn_algor,schwarzschild21_datas_study_gener_from_easy,bansal2022endtoend}.
\citep{graves16_adapt_comput_time_recur_neural_networ} and models that explicitly vary computation \citep{chung2024thinker}.
However, \Citet{knutson24_logic_extrap_mazes_with_recur_implic_networ} argue that they lack generalization and fail to
implement correct algorithms. Other works explore models that adjust computation per step, either with an explicit world model \citep{chung2024thinker} or without one \citep{graves16_adapt_comput_time_recur_neural_networ}. Our work suggests such adaptive strategies for long-term planning and reasoning can be learned implicitly using model-free RL, mirroring the recent advancements of LLMs leveraging test-time compute for improved reasoning \citep{deepseekai2025deepseekr1incentivizingreasoningcapability, chatgpto1}.

\paragraph{Goal misgeneralization and mesa-optimization.}
Understanding how neural networks perform tasks requiring sequential reasoning is crucial for ensuring transparency and safety of AI systems.
In AI alignment, goal misgeneralization occurs when a neural network optimizes for unintended proxy objectives instead of its intended goal (e.g.~\citealp{Yudkowsky2006ArtificialIA,omohundro-drives}; see \citealp{russell2019human}).
This risk relates to mesa-optimization, where a network develops internal objectives that diverge from training targets \citep{hubinger2019risks}. Recent work has shown that networks may pursue goals misaligned—or even directly opposed—to those intended by their designers \citep{di2022goal,shah2022goal}. Mechanistic interpretability of AI agents is a crucial tool for understanding how neural networks internally represent goals and plans and for identifying and fixing potential misalignments.

We discuss more related work  in \cref{app:more-related-work}.

\section{Conclusion}
\label{sec:conclusion}
Building on prior and concurrent work on search-like behaviors of DRC agents \citep{guez19model_free_planning} and plan representations \citep{tomsokoban}, we provide detailed evidence that DRC agents represent causal plans through spatial encodings of future box movements and refine them over time.

We describe how extra thinking steps increase success rate, by preventing greedy actions before the plan stabilizes into a longer-term solution. We find the agent takes advantage of this by ``pacing'' on-distribution to get enough time to stabilize the plan. By substituting the network output module with much simpler probes, we demonstrate how a mechanistic understanding can enable generalization beyond the agent's training distribution.

These findings advance the understanding of planning dynamics in neural networks and lay the groundwork for more transparent, safer and interpretable AI systems.

\ificlrfinal
\fi

\subsection*{Impact Statement}
This research into interpretability can make models more transparent, which helps in making models predictable, easier to debug and ensure they conform to specifications.

Specifically, we train and open-source a model organism which is planning, and analyze it somewhat; we hope this will catalyze further research on identifying, evaluating and understanding what \emph{goal} a model has.
We hope that directly identifying a model's goal lets us monitor for and correct goal misgeneralization \citep{di2022goal}.

Goal evaluation also has implications for AI welfare. Knowing the goals of a model would also make it possible to know whether these goals are being satisfied enough, and thus perhaps give a way to evaluate whether or not a model is happy (\cref{app:more-related-work}), assuming it is a moral patient.


\bibliography{main}

\appendix

\section{Training the test subject}
\label{app:training}

\paragraph{$\text{DRC}(D, N)$ architecture.}
\Citet{guez19model_free_planning} introduced the Deep Repeating ConvLSTM (DRC), whose core consists of $D$ convolutional
LSTM layers with 32 channels and $3 \times 3$ filters, each applied $N$ times per time step. Our \drcthree{} -- or just DRC for brevity -- has $1.29$M parameters. Before the LSTM core, two convolutional layers (without nonlinearity) encode the
observation with $4 \times 4$ filters.

The LSTM core uses $3\times 3$ convolutional filters, and a nonstandard \texttt{tanh} on the output gate \citep{jozefowicz15}. Unlike the original ConvLSTM \citep{convlstm}, the input to each layer of a DRC
consists of several concatenated components:
\begin{itemize}
\item The encoded observation is fed into each layer.
\item To allow spatial information to travel fast in the ConvLSTM layers, we apply \emph{pool-and-inject} by max- and mean-pooling the previous step's hidden state. We linearly combine these values channel-wise before feeding them as input to the next step.
\item To avoid convolution edge effects from disrupting the LSTM dynamics, we feed in a $12 \times 12$ channel
with zeros on the inside and ones on the boundary. Unlike the other inputs, this one is not zero-padded, maintaining the output size.
\end{itemize}

\paragraph{ResNet architecture.} This is a convolutional residual neural network, also from
\citet{guez19model_free_planning}. It serves as a non-recurrent baseline that can only think during the forward pass (no ability to think for extra steps) but is nevertheless good at the game. The ResNet consists of 9 blocks, each with $4 \times 4$ convolutional
filters. The first two blocks have 32 channels, and the others have 64. Each block consists of a convolution, followed by two (relu, conv) sub-blocks, each of which splits off and is added back to the trunk. The ResNet has $3.07$M parameters.

\paragraph{Value and policy heads.} After the convolutions, an affine layer projects the flattened spatial output
into 256 hidden units. We then apply a ReLU and two different affine layers: one for the actor (policy) and one for the
critic (value function).

\paragraph{RL training.} We train each network for 2.003 billion environment steps\footnote{A rounding error caused this to exceed $2\text{B}$ (\cref{sec:training-steps}).} using IMPALA
\citep{espeholt18_impal,huang2023cleanba}. For each training iteration, we collect 20 transitions on 256 actors using the network
parameters from the previous iteration, and simultaneously take a gradient step. We use a discount rate of $\gamma=0.97$ and
V-trace $\lambda=0.5$. The value and entropy loss coefficients are $0.25$ and $0.01$. We use the Adam
optimizer with a learning rate of $4 \cdot 10^{-4}$, which linearly anneals to $4 \cdot 10^{-6}$ at the end of training. We clip the gradient norm to $2.5 \cdot 10^{-4}$. Our hyperparameters
are mostly the same as \citet{guez19model_free_planning}; see \cref{app:hyperparameters}.

\paragraph{A* solver.} We used the A* search algorithm to obtain optimal solutions to each Sokoban puzzle. The heuristic was the sum of the Manhattan distances of each box to its nearest target. Solving a single level on one CPU
takes anywhere from a few seconds to 15 minutes.\footnote{The A* solutions may be of independent interest, so we make them available at \url{https://huggingface.co/datasets/AlignmentResearch/boxoban-astar-solutions/}.}

\subsection{Training hyperparameters}
\label{app:hyperparameters}

All networks were trained with the same hyperparameters, tuned on both ResNet and the \drcthree{}. These largely match \citet{guez19model_free_planning}, except we take the \emph{mean} per-step losses instead of summing.

\paragraph{Time limits.} During training, we want to prevent strong time correlations between the returns, so the
gradient steps are not correlated over time. For this reason, the time limit for each episode is uniformly random
between 91 and 120 time steps.

\paragraph{Loss.} The value and entropy coefficients are 0.25 and 0.01 respectively. It is very important to \emph{not} normalize the advantages for the policy gradient step.

\paragraph{Gradient clipping and epsilon} The original IMPALA implementation, as well as \citet{huang2023cleanba}, \emph{sum} the per-step losses. We instead average them for more predictability across batch sizes, so we had to scale down some parameters by a factor of $1/640$: Adam $\epsilon$, gradient norm for clipping, and L2 regularization).

\paragraph{Weight initialization.} We initialize the network with the Flax \citep{flax2020github} default: normal weights truncated at 2 standard deviations and scaled to have standard deviation $\sqrt{1/\text{fan\_in}}$. Biases are initialized to 0. The forget gate of LSTMs has 1 added to it \citep{jozefowicz15}. We initialize the value and policy head weights with orthogonal vectors of norm 1.
Surprisingly, this makes the variance of these unnormalized residual networks decently close to 1.

\paragraph{Adam optimizer.} As our batch size is medium-sized, we pick $\beta_1=0.9$, $\beta_2=0.99$. The denominator epsilon is $\epsilon = 1.5625 \cdot 10^{-7}$. Learning rate anneals from $4\cdot 10^{-4}$ at the beginning to $4\cdot 10^{-6}$ at 2,002,944,000 steps.

\paragraph{L2 regularization.} In the training loss, we regularize the policy logits with L2 regularization with coefficient $1.5625 \times 10^{-6}$. We regularize the actor and critic heads' weights with L2 at coefficient $1.5625 \times 10^{-8}$. We believe this has essentially no effect, but we left it in to more closely match \citet{guez19model_free_planning}.

\paragraph{Software.} We base our IMPALA implementation on Cleanba \citep{huang2023cleanba}. We implemented Sokoban in C++ using Envpool \citep{weng2022envpool} for faster training, based on gym-sokoban \citep{SchraderSokoban2018}.

\subsection{Number of training steps}
\label{sec:training-steps}
The paper states networks train for $2.003\text{B}$ steps, but the exact number is $2\,002\,944\,000$ steps. Our code and hyperparameters require that the number of environment steps be divisible by $5\,120 =  256\text{ environments}\times 20\text{ steps collected}$, because that is the number of steps in one iteration of data collection.

However, $2\text{B}$ is divisible by $5\,120$, so there is no need to add a remainder. We noticed this mistake once the networks already have trained. Retraining the networks to correct this error was deemed unnecessary.

At some point in development, we settled on $80\,025\,600$ to approximate $80\text{M}$ while being divisible by $256\times 20$ and $192 \times 20$. Perhaps due to error, this mutated into $1\,001\,472\,000$ as an approximation to $1\text{B}$, which directly leads to the number we used.

\subsection{Learning curve comparison}
\label{app:replication}
Replicating the results of \citet{guez19model_free_planning} proved challenging.  \citet{chung2024thinker} propose an improved method for RL in planning-heavy domains. They employ the IMPALA \drcthree{} as a baseline and plot its performance in \citet[Figure~5]{chung2024thinker}. They plot two separate curves for \drcthree{}: that from \citet{guez19model_free_planning}, and a decent replicated baseline. The baseline is considerably slower to learn and peaks at lower performance.

We did not innovate in RL, so were able to spend more time on the replication.
We compare our replication to \citet{guez19model_free_planning} in \cref{fig:test-valid-learning-curves}, which shows that the learning curves for \drcthree{} and ResNet are compatible, but not the one for $\drcone{}$. Our implementation exhibits reduced stablility, with large error bars and pronounced oscillations over time. We defer addressing this to future work. \Cref{tab:unfil_test_metrics} reports test and validation performance for the DRC and ResNet seeds which we picked for the paper body.

The parameter counts (\cref{tab:parameter-counts}) are very different from what \citet{guez19model_free_planning} report. In private communication with the authors, we confirmed that our architecture has a comparable number of parameters, and some of the originally reported numbers are a typographical error.

\begin{tabhalfwidth}[tp]{Parameter counts for each architecture.}
\label{tab:parameter-counts}
\begin{tabular}{lrr}
\toprule
Architecture & Parameter count & {} \\
\midrule
\drcthree{} & 1,285,125 &(1.29M)\\
\drcone{} &  987,525 &(0.99M)\\
ResNet: & 3,068,421 &(3.07M) \\
\bottomrule
\end{tabular}
\end{tabhalfwidth}

\begin{figure}[tbp]
    \centering
    \includegraphics[width=\linewidth]{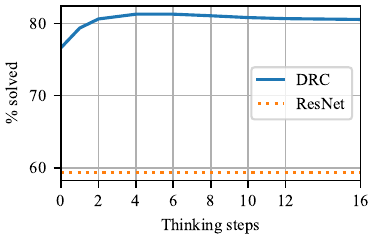}
    \caption{Success rate of \drcthree{} on Validation-medium levels across forced thinking steps at the start of the episode. Increasing thinking steps increases performance.}
    \label{fig:success_rate_with_thinking}
\end{figure}


\begin{figfullwidth}[hbtp]{Success rate for Test-unfiltered and Validation-medium levels vs. environment steps of training. Each architecture has 5 random seeds, the solid line is the pointwise median and the shaded area spans from the minimum to the maximum. The dotted lines are data for the performance of architectures extracted from the \citep{guez19model_free_planning} PDF file.}
\includegraphics[width=\linewidth]{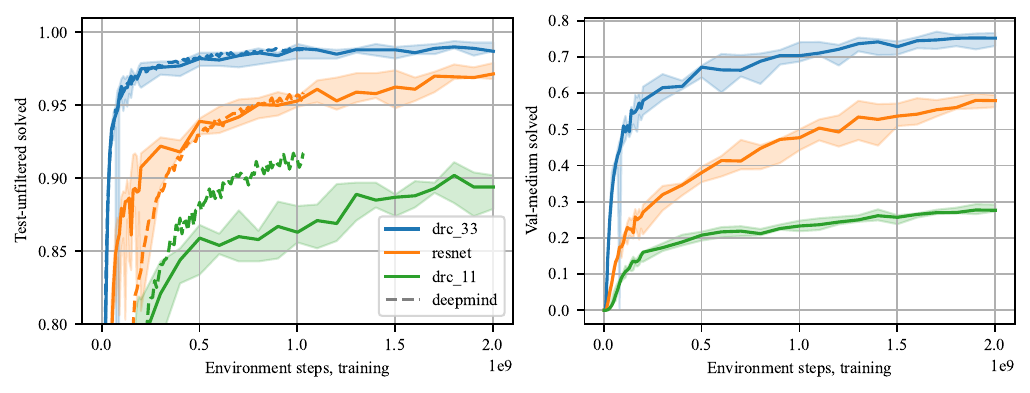}
\label{fig:test-valid-learning-curves}
\end{figfullwidth}

\begin{tabfullwidth}[ht]{Success rate and return of DRC and ResNet on the unfiltered test set at various training environment steps.\label{tab:unfil_test_metrics}}
\begin{tabular}{ccccccccc}
\toprule
\textsc{Training Env} & \multicolumn{4}{c}{\textsc{Test Unfiltered}} & \multicolumn{4}{c}{\textsc{Valid Medium}} \\
\textsc{Steps} & \multicolumn{2}{c}{\textsc{ResNet}} & \multicolumn{2}{c}{\textsc{DRC}(3, 3)} & \multicolumn{2}{c}{\textsc{ResNet}} & \multicolumn{2}{c}{\textsc{DRC}(3, 3)} \\
& \textsc{Success} & \textsc{Return} & \textsc{Success} & \textsc{Return} & \textsc{Success} & \textsc{Return} & \textsc{Success} & \textsc{Return} \\ \midrule
100M & 87.8 & 8.13 & 95.4 & 9.58 & 18.6 & -6.59 & 47.9 & -0.98 \\
500M & 93.1 & 9.24 & 97.9 & 10.21 & 39.7 & -2.64 & 66.6 & 2.62\\
1B & 95.4 & 9.75 & 99.2 & 10.47 & 50.0 & -0.64 & 70.4 & 3.40 \\
2B & 97.9 & 10.29 & 99.3 & 10.52 & 59.4 & 1.16 & 76.6 & 4.52 \\
\bottomrule
\end{tabular}
\end{tabfullwidth}

\section{Results on \drcone{} and \resnet{}}\label{app:results_on_drc11_and_resnet}

\subsection{\drcone{}}
In this section, we show that the \drcone{} network, which only has a single recurrent convolutional block that runs a single forward pass per environment step, also exhibits similar planning behavior to the \drcthree{} network, although achieving lower performance as expected due to lower parameter count and compute expenditure per environment step. The parameter count for the \drcone{} and other networks used in the paper are reported in \cref{tab:parameter-counts}.

\begin{figure}
    \centering
    \includegraphics[width=0.9\linewidth]{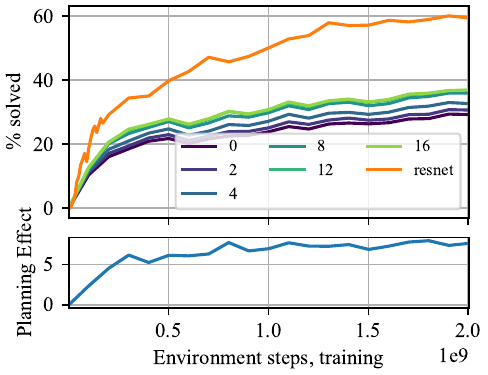}
    \caption{The \drcone{} network shows similar planning effect as the \drcthree{}. The plot is computed on the medium difficulty validation levels. The planning effect is measured as the difference in performance between the best thinking step and no thinking step. It achieves a lower performance than the $9$ layer ResNet network due to lower parameter count.}
    \label{fig:drc11_headline_figure}
\end{figure}

\Cref{fig:drc11_headline_figure} shows the performance of the \drcone{} network on the medium difficulty validation levels during training. The \drcone{} network shows a similar planning effect as the \drcthree{} network (\cref{fig:headline-figure}), which matches the results of \citet{guez19model_free_planning} who also showed that extra thinking steps improved performance of both the \drcone{} and the \drcthree{} network.

We train the box-directions probe on the \drcone{} network and evaluate it on the medium validation levels. The resulting probe achieves an F1 score of $72.3\%$. The F1 score is high enough to suggest that the \drcone{} network also has a plan representation similar to the \drcthree{} network. Upon inspecting the probe visualizations, we find that the probe is indeed good at predicting the box movements in the short-term but is worse than the probe on \drcthree{} at predicting the long-term box movements (\cref{fig:drc11_chain}, left). This suggests that the \drcone{} network has a plan-representation, but it is limited to short-term planning. This results matches the evidence from the network's overall performance which is significantly worse on level requiring long-term planning.

We now ask whether the performance improvement during thinking steps can be explained by the network refining its plan as we showed for the \drcthree{} network in \cref{sec:plan-improves}. We measure the plan quality as we did in the main text using the length of plan from boxes and number of positive predictions. We find that the plan quality (\cref{fig:drc11_chain}, middle) as well as the F1 score (\cref{fig:drc11_chain}, right) measured using the probe increases over the number of thinking steps. \Cref{fig:drc11_time_to_box_and_opt_length} (left) shows that this network also avoids myopic strategies when given extra thinking steps. This shows that the \drcone{} network also refines its plan during the thinking steps.

\begin{figure}[t]
    \centering
    \includegraphics[width=\linewidth]{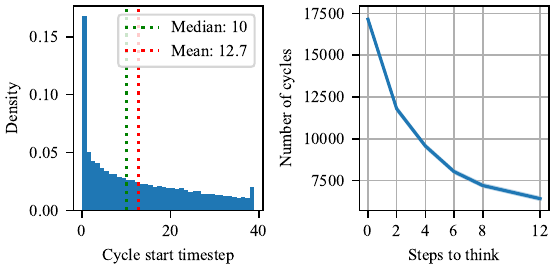}
    \caption{\textbf{Left:} \drcone{} also performs cycles at the start of levels when it is more beneficial. \textbf{Right:} \drcone{} performs many cycles when given no thinking steps. More than 60\% of the cycles disappear with extra thinking steps.}
    \label{fig:drc11_cycles}
\end{figure}

Finally, we show that the \drcone{} network also exhibits the pacing behavior as the \drcthree{} network. \Cref{fig:drc11_cycles} (left) shows that \drcone{} also performs cycles at the start of levels, where extra computation steps are more useful to come up with better solutions. \Cref{fig:drc11_cycles} (right) shows that 60\% of these cycles disappear with extra thinking steps, suggesting that they are performed by deliberately and not accidentally. The percentage is lower than 75\% for \drcthree{} suggesting that \drcone{} performs comparatively more accidental cycles, which is again consistent with lower planning-capacity and performance of the network.

Hence, the main Findings~\ref{hyp:plan}$-$\ref{hyp:pacing} analysed for \drcthree{} in the main text are also true for \drcone{}, with results suggesting worse long-term planning as compared to \drcthree{} as expected.

\begin{figfullwidth}[t]{\textbf{Left:} Thinking steps also help \drcone{} to avoid myopic plans, thus increasing the time to place boxes first 3 boxes in solved levels while solving the level faster by placing the box 4 earlier. \textbf{Right:} The average optimal length of level solved by extra thinking steps doesn't change with more thinking steps. This suggests that, unlike \drcthree{} which solves more difficult levels with more thinking steps, \drcone{}'s planning capacity is limited. \label{fig:drc11_time_to_box_and_opt_length}}
\includegraphics[width=\linewidth]{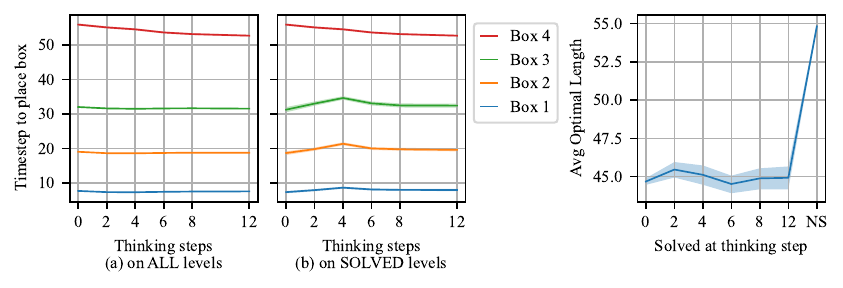}
\end{figfullwidth}

\begin{figfullwidth}[ht]{\textbf{Left:} For \drcone{}, the number of steps in advance that the Box-Directions probe is able to predict a box move shows similar trend as \drcthree{}, but is overall lower, indicating that the \drcone{} is less capable of planning. \textbf{Middle and Right:} The plan quality, as measured by summing probe-predicted chain lengths starting from boxes and counting squares with positive non-empty predictions, and the F1-score increases over thinking steps (\cref{sec:probes-and-steering}). This suggests that \drcone{} also refines its plan during computation, although it has an overall lesser F1 score of $60\%$ as compared to $84\%$ for \drcthree{}.} \label{fig:drc11_chain}
\includegraphics[width=0.33\linewidth]{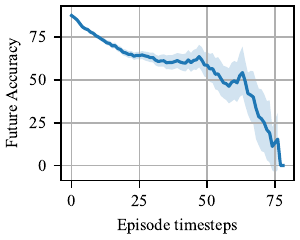}
\includegraphics[width=0.33\linewidth]{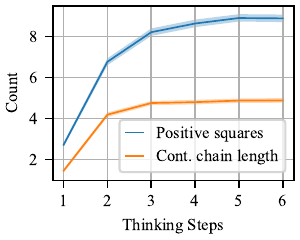}
\includegraphics[width=0.33\linewidth]{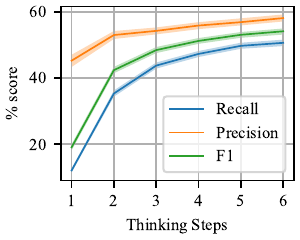}
\end{figfullwidth}

\begin{figfullwidth}[htp]{\textbf{Left:} The box-directions probe trained on all layers for the \resnet{} model can predict moves with $80\%$ accuracy 50 steps before they occur. This implies that the \resnet{} model also computes a long-term plan for every observation. \textbf{Middle:} The plan quality and accuracy measured with the all-layer box-directions probe doesn't improve with more environment steps as expected, since the plan in the activations need to be recomputed entirely from scratch on every step. This is unlike the DRC models which improve their plans with more thinking and environment steps. \textbf{Right:} Training the box-directions probe on each layer separately reveals that the plan representation is computed and improved in the early middle layers 1-4 with F1-score reaching 63.9\%. The layer likely compute different parts of the plan since the probe trained on all the layers achieves a significantly higher F1-score of 84.5\%. \label{fig:resnet_chain}}
\includegraphics[width=0.245\linewidth]{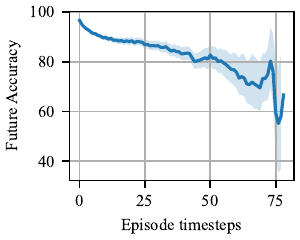}
\includegraphics[width=0.245\linewidth]{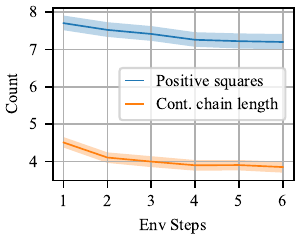}
\includegraphics[width=0.245\linewidth]{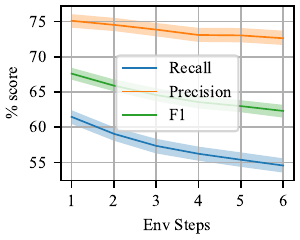}
\includegraphics[width=0.245\linewidth]{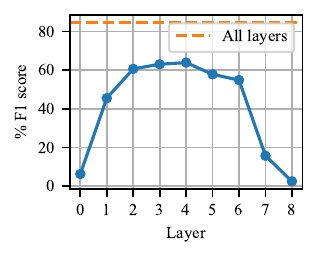}
\end{figfullwidth}

\subsection{\resnet{}}

We now check if our findings about the DRC networks also generalize to a \resnet{} model with no recurrence over hidden state. The training and architecture details of the \resnet{} model is provided in \cref{app:training}.

We first train the box-directions probe on the concatenation of the output of the ReLU activations from every residual block in every layer. This creates a single input vector of size $1024$ for the linear probe. The activations are collected on 1000 uniformly sampled levels from the medium-difficulty train set on which the \resnet{} model has a solve rate of $59.9\%$. The resulting probe achieves an F1 score of $84.5\%$ for predicting future box-movement directions on the validation set, similar to the probe on the \drcthree{} network. \Cref{fig:resnet_chain} (left) also shows that the activations of the model in early steps are highly predictive of the actions it will take many steps in the future. This suggests that both the networks have similar representations that store their long-term plans. However, since the \resnet{} model achieves a lower performance on medium and hard difficulty sets, we can conclude that the \resnet{} model is not as good at constructing correct plans as the \drcthree{} network.

We found that plan improves with more thinking steps for the DRC model. Since the \resnet{} is not a recurrent model, it cannot take in extra thinking steps like the DRC networks can. Since the \resnet{} does not have a recurrent hidden state, we would expect that the plan representation of the model should not improve with more environments steps, as the activations are recomputed from scratch on every step. \Cref{fig:resnet_chain} (middle plots) show that the plan is computed in one-step, and more environment steps do not result in better plan representations. \Cref{fig:resnet_chain} (right) shows the validation F1-score of the same probe trained on the activations of the individual layers separately. We can see that the plan improves in the early to middle layers, implying that the plan representations are sequentially through the layers. Since the \resnet{} is state-less, it cannot exhibit the pacing-behavior.

Hence, we can conclude that only the \cref{hyp:plan} of plan representation holds for the \resnet{} model, with plan-improvements happening sequentially through the layers.

\section{Generalizing the \texorpdfstring{\drcthree{}}{DRC(3, 3)} to larger levels}
\label{app:borgar-levels}

We license the levels by \citet{borgar-sokoban} as GPLv3, and make them available in the supplementary material. \Cref{tab:performance-big-levels} shows the performance of DRC acriss various level sets. Higher
scores align with levels considered easier by humans; for example, ``Dimitri \& Yorick'' was made for children by Jacques
Duthen, features small levels (maximum size: $12\times 10$) with at most 5 boxes. Level sets where
the DRC fails to solve any puzzles were also challenging for the authors.

We tested the \drcthree{} with 2 to 128 extra thinking steps, incrementing in powers of two. For
most sets, we find some benefit to 2-4 thinking steps, but no more. The sole exception is XSokoban, which contains one level (\cref{fig:successful-larger-levels}(c)) requiring 128 steps of thinking to solve.

We encourage the reader to go to the website by \citet{borgar-sokoban} and try solving the levels.

\begin{figfullwidth}[hbtp]{Thinking steps are not always useful: For level (a), the network succeeds with 0 to 16 thinking steps, but fails with 32 or more thinking steps. Levels (b) and (c) are too difficult for the \drcthree{}, and it always fails on them independent of thinking steps, though it comes up with decent partial solutions. All levels from \citet{borgar-sokoban}.}
\begin{subfigure}{0.24\linewidth}%
\includegraphics[width=\linewidth]{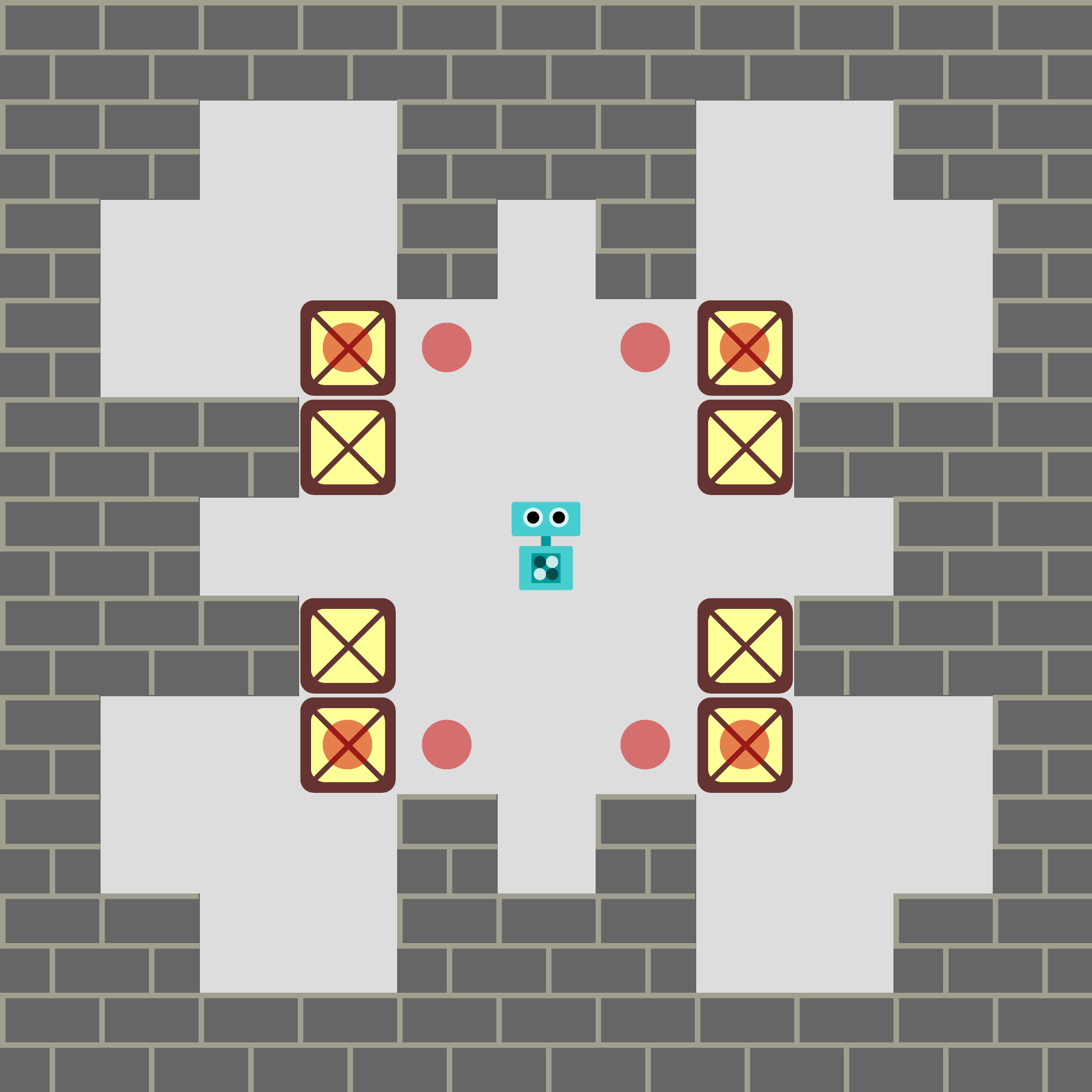}%
\caption{Microban, level 105}%
\end{subfigure}%
\hspace{0.01\linewidth}%
\begin{subfigure}{0.24\linewidth}%
\includegraphics[width=\linewidth]{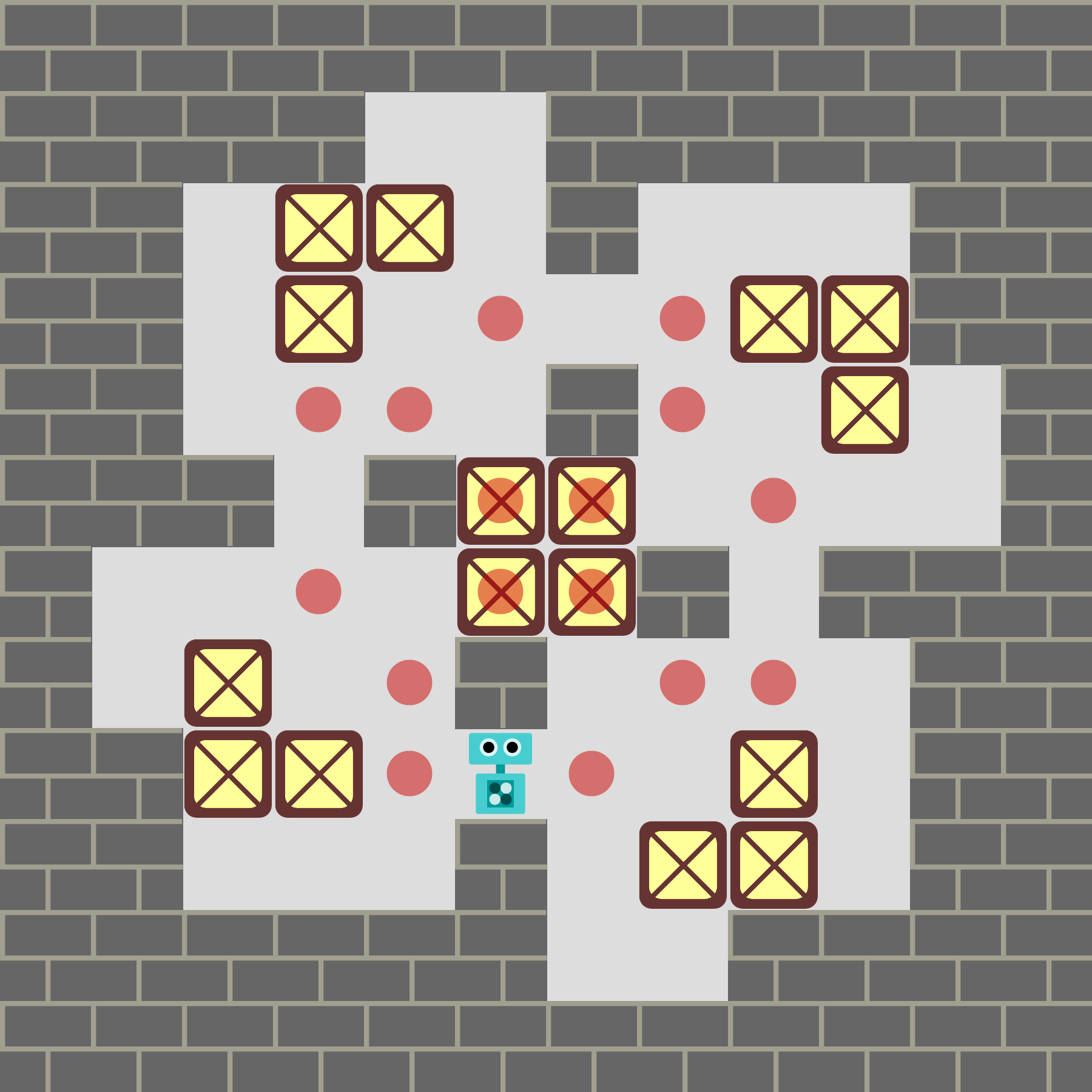}%
\caption{Microban, level 144}%
\end{subfigure}%
\hspace{0.01\linewidth}%
\begin{subfigure}{0.5\linewidth}%
\includegraphics[width=\linewidth]{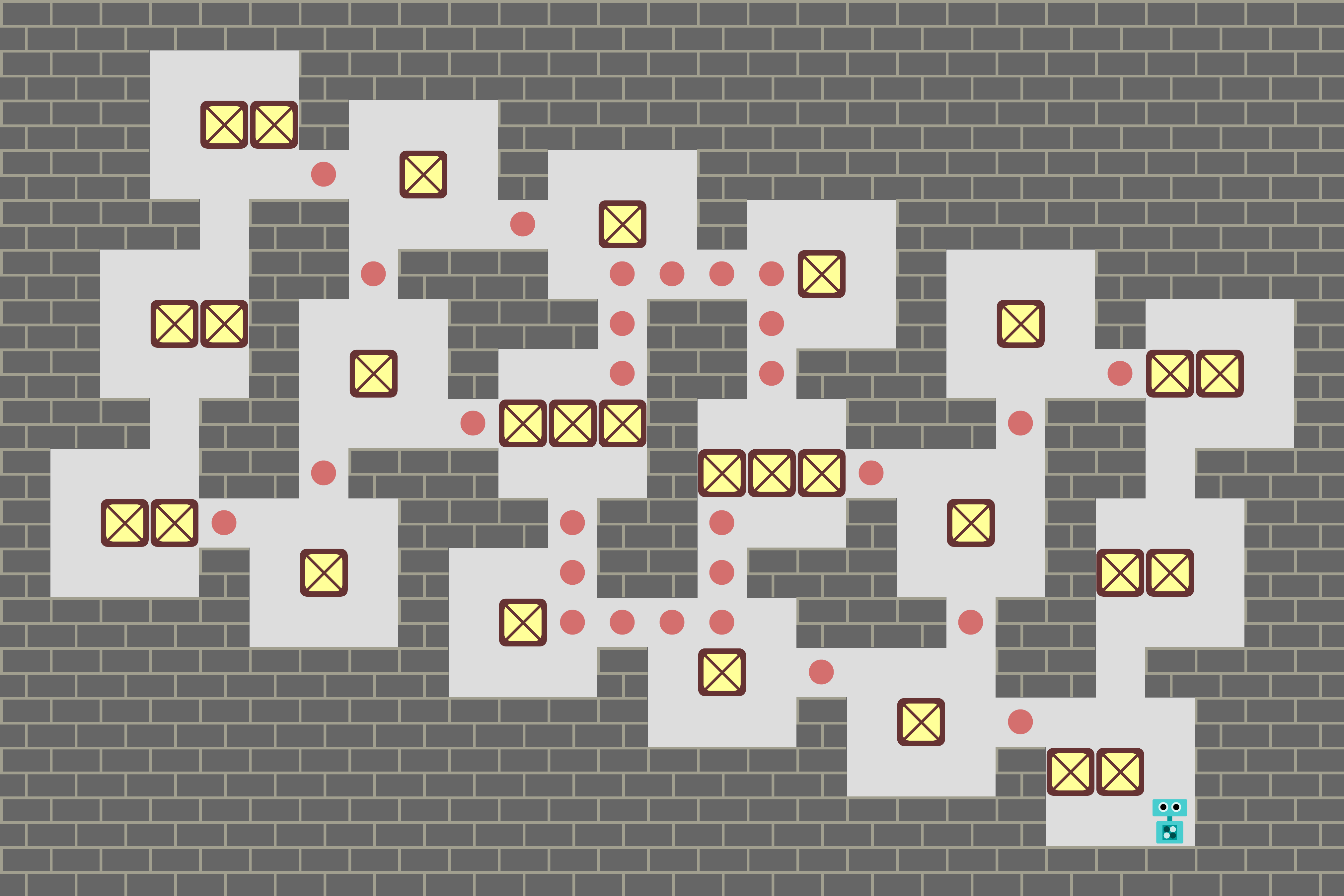}%
\caption{Mas Sasquatch, level 15}%
\end{subfigure}%
\end{figfullwidth}

\begin{figfullwidth}[ht]{Some levels require thinking steps for success: For level (a), the network always succeeds. For level (b), it succeeds at 32 or more thinking steps. For level (c), it needs 128 thinking steps to succeed. All levels from \citet{borgar-sokoban}.\label{fig:successful-larger-levels}}
\begin{subfigure}{0.28\linewidth}%
\includegraphics[width=\linewidth]{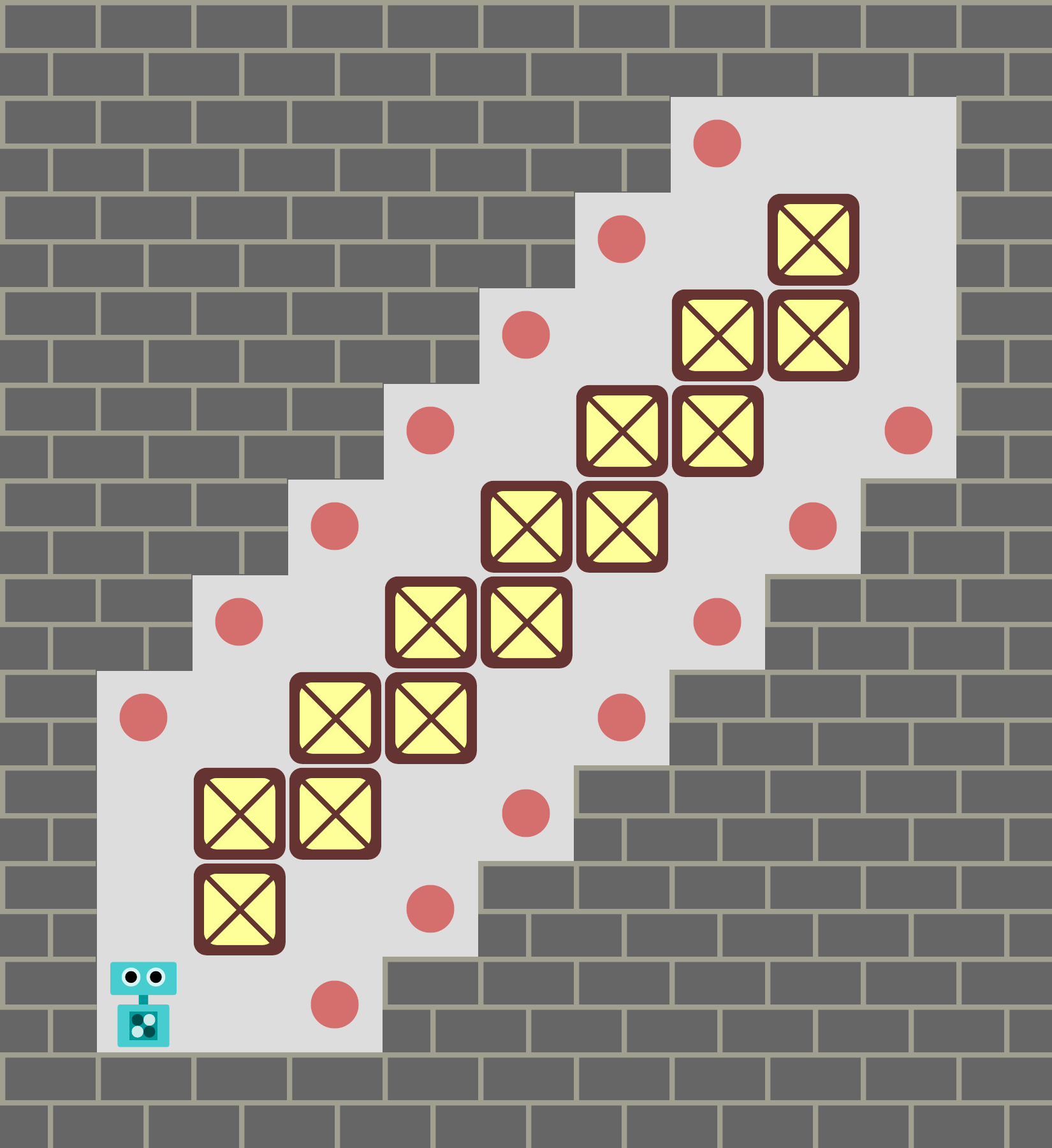}%
\caption{Sokoban Jr. 2, level 17}%
\end{subfigure}%
\hspace{0.01\linewidth}%
\begin{subfigure}{0.28\linewidth}%
\includegraphics[width=\linewidth]{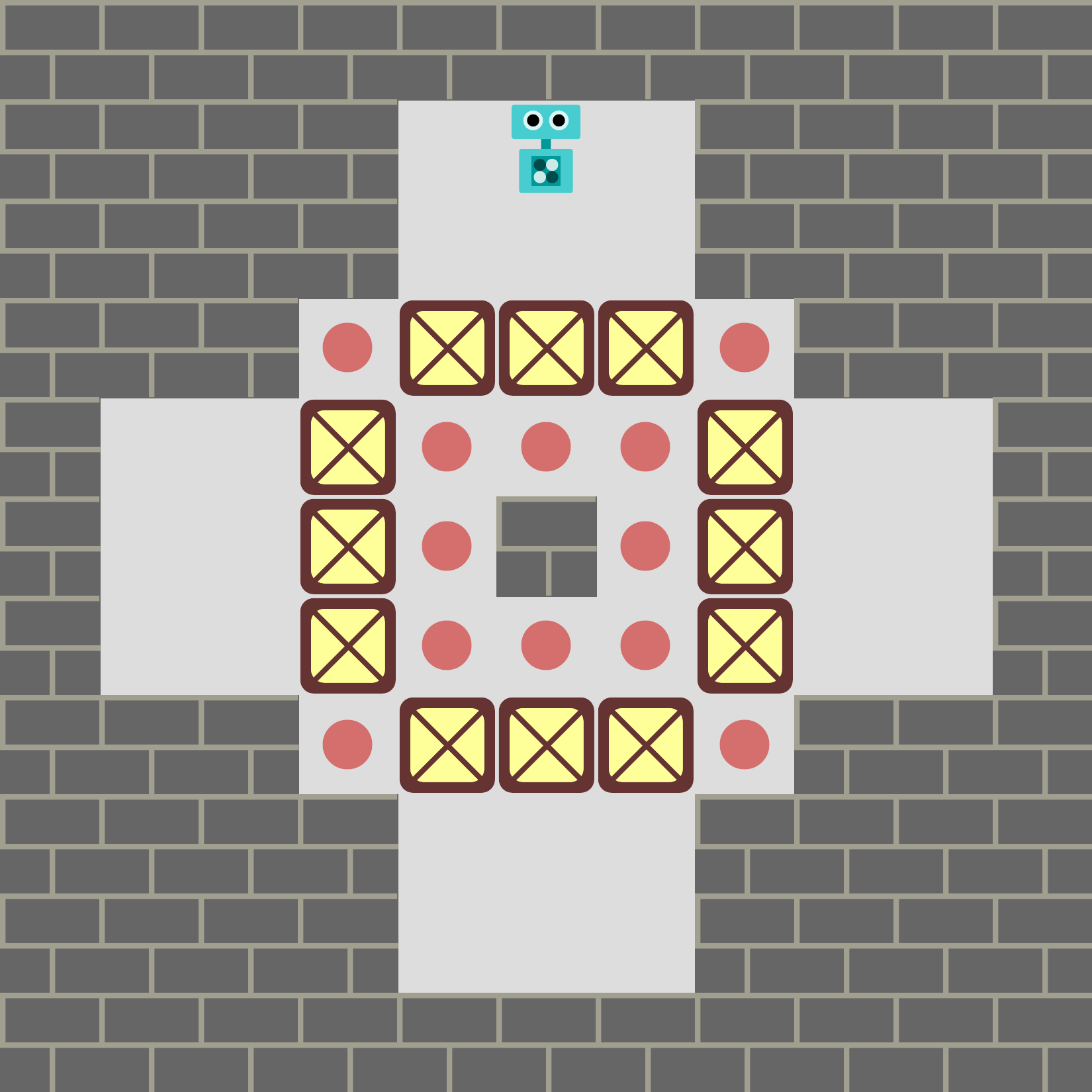}%
\caption{Microban, level 145}%
\end{subfigure}%
\hspace{0.01\linewidth}%
\begin{subfigure}{0.42\linewidth}%
\includegraphics[width=\linewidth]{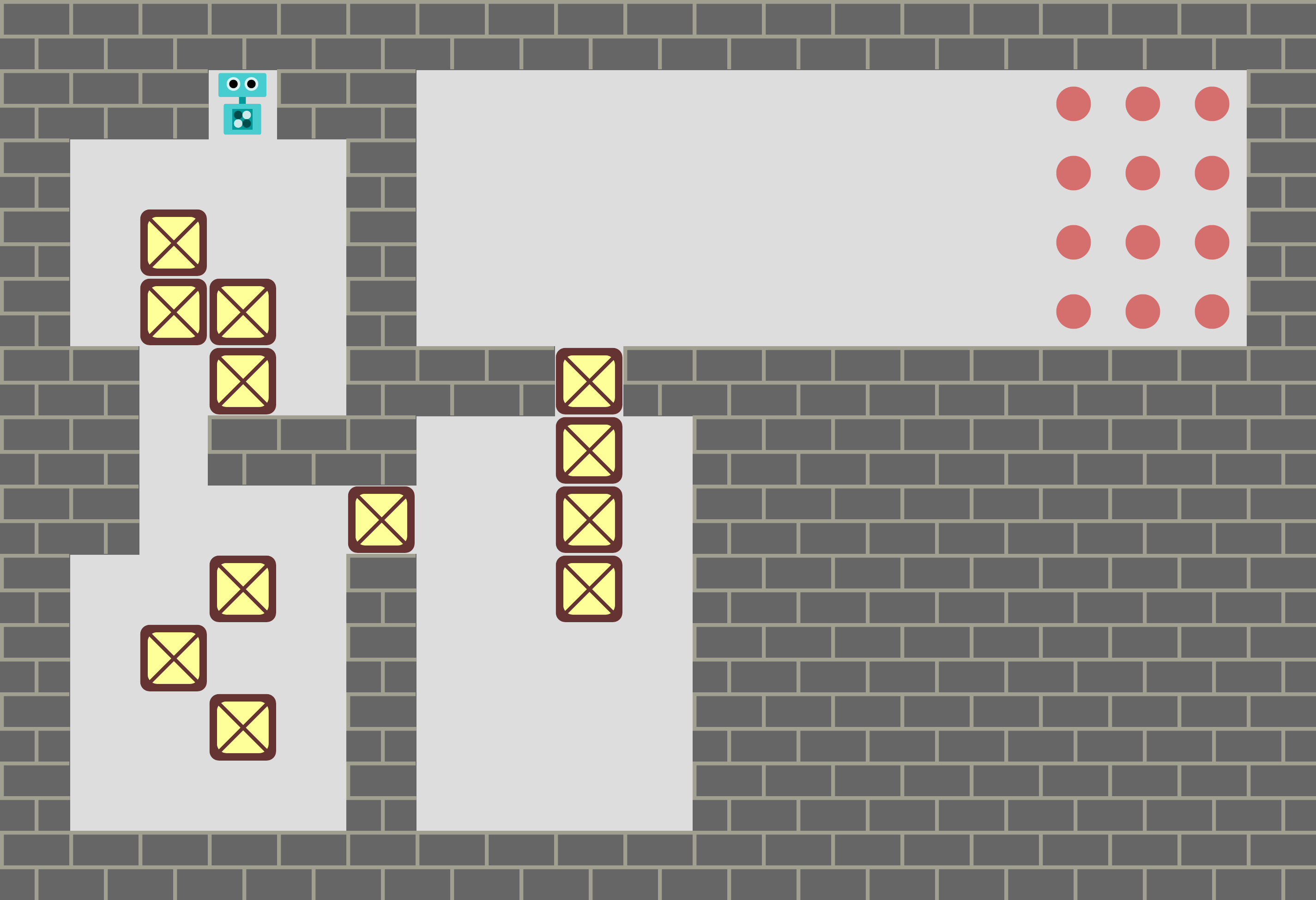}\label{fig:xsokoban-31}%
\caption{XSokoban, level 30}%
\end{subfigure}%
\end{figfullwidth}

\begin{tabfullwidth}[hbtp]{Performance of the \drcthree{} on each set of levels by \citep{borgar-sokoban}. The ``max solved'' columns
  represent the proportion of levels solved at the number of steps in the ``max at'' column, which is the highest solved proportion for each number of thinking steps tried. The ``largest level size'' column represents the height and width of the solved level with the largest grid area.\label{tab:performance-big-levels}}
\begin{tabular}{l|rlll|rlll|l}
\toprule
& \multicolumn{4}{|c|}{\textsc{All levels}} & \multicolumn{4}{|c|}{\textsc{Levels larger than $10\times 10$}} &
\multicolumn{1}{c}{\textsc{Largest}} \\
\textsc{Level collection} &
\textsc{\#} & \textsc{solved} & \textsc{Max slv.} & \textsc{max at} &
\textsc{\#} & \textsc{solved} & \textsc{Max slv.} & \textsc{max at} & \multicolumn{1}{c}{\textsc{level size}} \\
\midrule

Dimitri \& Yorick & 61 & 86.9\% & 93.4\% & 2 & 0 & --- & --- & --- & (10, 12) \\
Sokoban Jr. 1 & 60 & 85.0\% & 85.0\% & 0 & 19 & 73.7\% & 73.7\% & 0 & (18, 21) \\
Howard's 3rd set & 40 & 70.0\% & 70.0\% & 0 & 1 & 0.0\% & 0.0\% & 0 & (11, 10) \\
Simple sokoban & 61 & 55.7\% & 62.3\% & 16 & 51 & 47.1\% & 54.9\% & 16 & (16, 19) \\
Sokoban Jr. 2 & 54 & 48.1\% & 53.7\% & 2 & 40 & 45.0\% & 47.5\% & 2 & (14, 27) \\
Sokogen 990602 & 78 & 37.2\% & 43.6\% & 4 & 0 & --- & --- & --- & (10, 10) \\
Microban & 155 & 31.0\% & 31.6\% & 64 & 17 & 5.9\% & 11.8\% & 2 & (14, 10) \\
Yoshio Automatic & 52 & 28.8\% & 36.5\% & 2 & 0 & --- & --- & --- & (10, 10) \\
Deluxe & 55 & 25.5\% & 27.3\% & 16 & 1 & 0.0\% & 0.0\% & 0 & (10, 13) \\
Howard's 2nd set & 40 & 12.5\% & 15.0\% & 2 & 22 & 0.0\% & 0.0\% & 0 & (11, 10) \\
Sasquatch III & 16 & 6.2\% & 6.2\% & 0 & 8 & 0.0\% & 0.0\% & 0 & (10, 17) \\
Microcosmos & 40 & 5.0\% & 10.0\% & 16 & 0 & --- & --- & --- & (10, 10) \\
Howard's 1st set & 100 & 4.0\% & 4.0\% & 0 & 54 & 0.0\% & 0.0\% & 0 & (12, 10) \\
Still more levels & 35 & 2.9\% & 2.9\% & 0 & 34 & 2.9\% & 2.9\% & 0 & (13, 13) \\
Sasquatch IV & 36 & 2.8\% & 2.8\% & 0 & 20 & 0.0\% & 0.0\% & 0 & (10, 10) \\
Xsokoban & 40 & 2.5\% & 5.0\% & 128 & 39 & 2.6\% & 5.1\% & 128 & (13, 19) \\
Sasquatch & 49 & 2.0\% & 4.1\% & 4 & 39 & 0.0\% & 2.6\% & 4 & (14, 24) \\
David Holland 1 & 10 & 0.0\% & 0.0\% & 0 & 5 & 0.0\% & 0.0\% & 0 & --- \\
David Holland 2 & 10 & 0.0\% & 0.0\% & 0 & 9 & 0.0\% & 0.0\% & 0 & --- \\
Howard's 4th set & 32 & 0.0\% & 0.0\% & 0 & 30 & 0.0\% & 0.0\% & 0 & --- \\
Mas Sasquatch & 50 & 0.0\% & 0.0\% & 0 & 43 & 0.0\% & 0.0\% & 0 & --- \\
Nabokosmos & 40 & 0.0\% & 0.0\% & 0 & 0 & --- & --- & --- & --- \\
Sokoban & 50 & 0.0\% & 0.0\% & 0 & 48 & 0.0\% & 0.0\% & 0 & --- \\
\bottomrule
\end{tabular}
\end{tabfullwidth}

\subsection{Generalization to zig-zag levels}

\begin{figure}[htbp]
    \centering
    \includegraphics[width=0.9\linewidth]{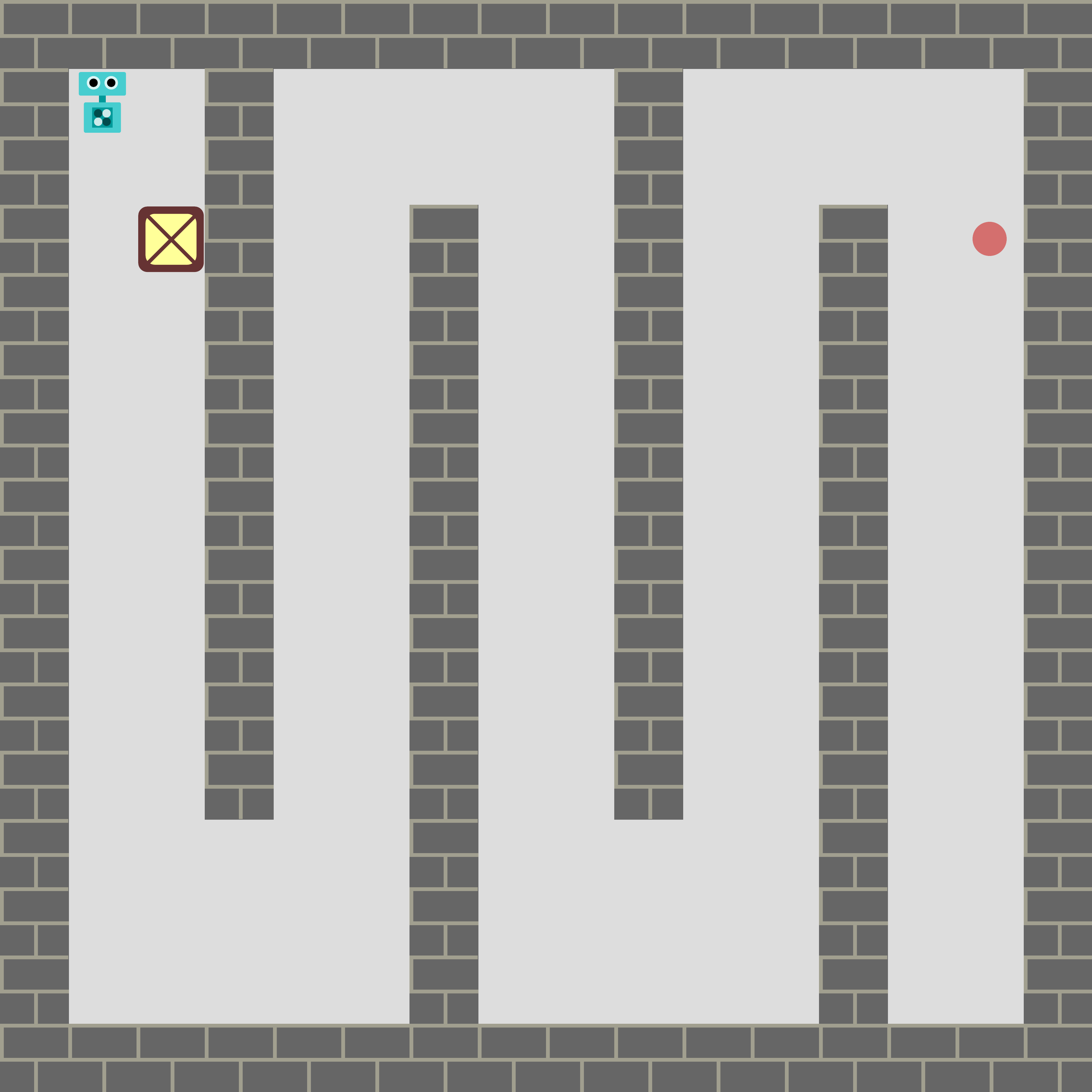}
    \caption{$16 \times 16$ zig-zag level that the \drcthree{} doesn't solve. The network solves all levels of this pattern only up to $15 \times 15$ size and fails on all zig-zag levels beyond that.}
    \label{fig:zig-zag-level}
\end{figure}

We create a template of levels as shown in \cref{fig:zig-zag-level} where the agent has to take boxes from the left side to the targets on the right side by going through multiple zig-zag vertical lanes. This template can be created for arbitrary sized levels by adjusting the number of vertical lanes that the agent has to pass through. We find that the agent is able to solve levels only until the size of $15 \times 15$ with 5 lanes but fails to solve larger levels. In larger levels, the agent does push the boxes through some of the lanes but is unable to find the complete path to the target through the remaining zig-zag lanes.

\section{Bistable and unstable plans in toy environments}\label{app:toy-levels}
\begin{figfullwidth}[htp]{\textbf{Left:} A custom level where the agent has two equally good paths to follow. The arrows show the prediction of the Agent-Directions probe with opacity proportional to the number of times an arrow was predicted
      across all the steps. The green and red arrows are correct and incorrect predictions, respectively. The agent behaves
      suboptimally by returning to the start and going left after first going right for three steps. \textbf{Middle:}
      The agent takes the path on the right after intervening with the corresponding arrows using the Agent-Directions
      probe on the first step. The same happens on the left if we intervene on that path without the suboptimal steps
      right of the undisturbed agent. \textbf{Right:} An empty level with Box-Directions probe intervening on the first
      four steps which are followed correctly by the agent on those four steps. When the intervention is removed on the
      fifth step, the agent computes the simpler path in green and doesn't follow the path laid out earlier (in red).
    \label{fig:toy-levels-intervention}}
    \begin{minipage}[t]{0.31\linewidth}
      \includegraphics[width=\linewidth]{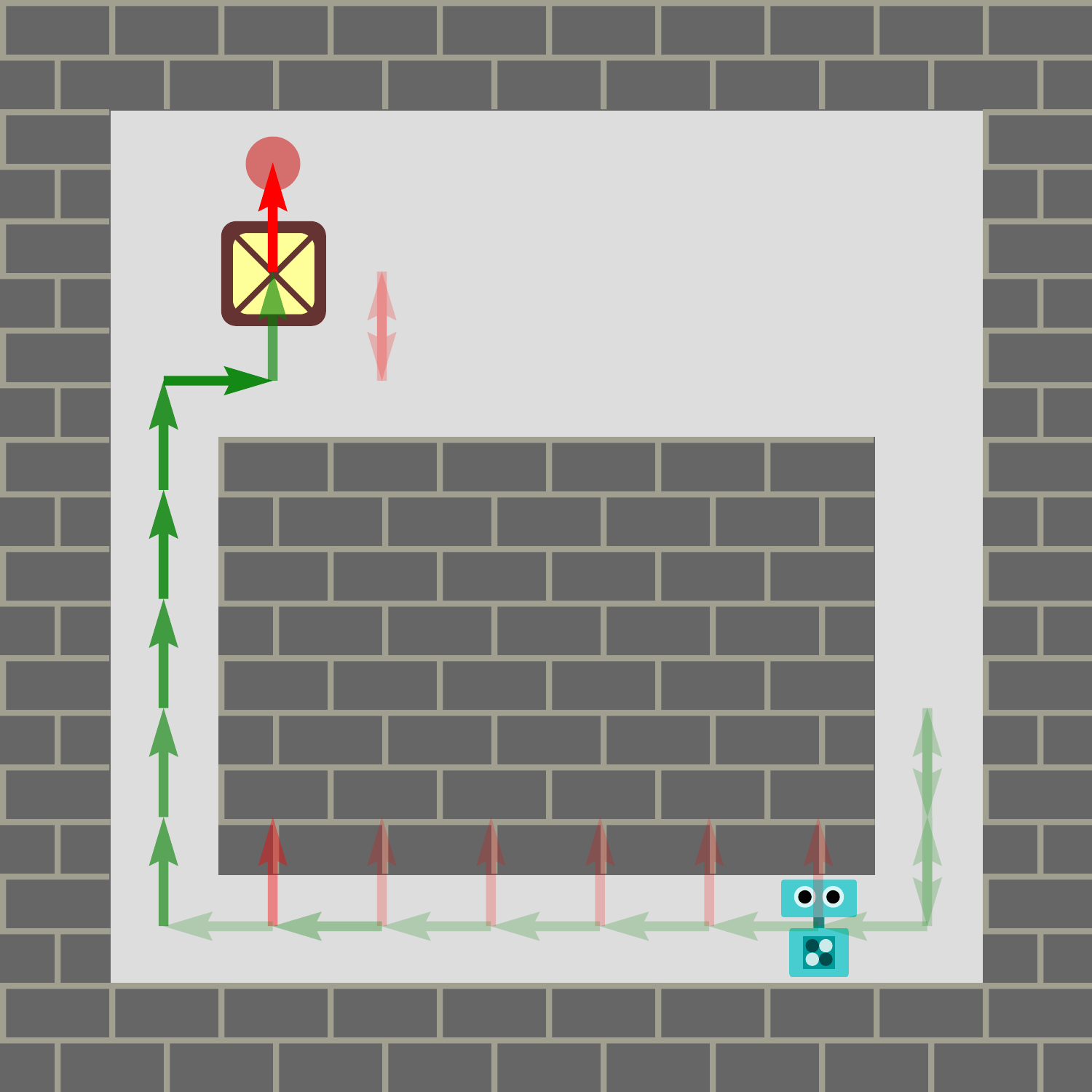}%
    \end{minipage}%
    \begin{minipage}[t]{0.31\linewidth}
      \includegraphics[width=\linewidth]{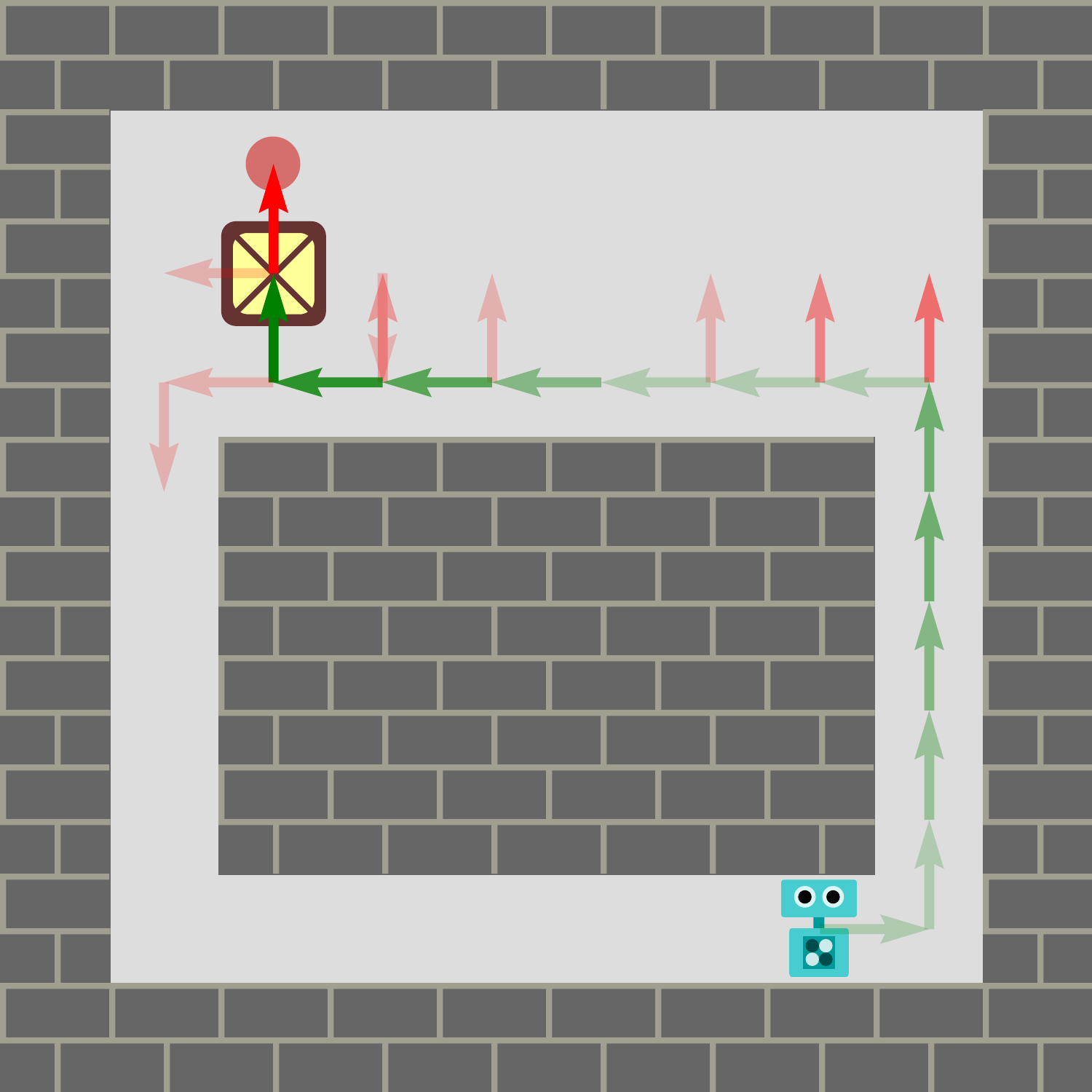}%
    \end{minipage}%
    \begin{minipage}[t]{0.31\linewidth}
        \includegraphics[width=\linewidth]{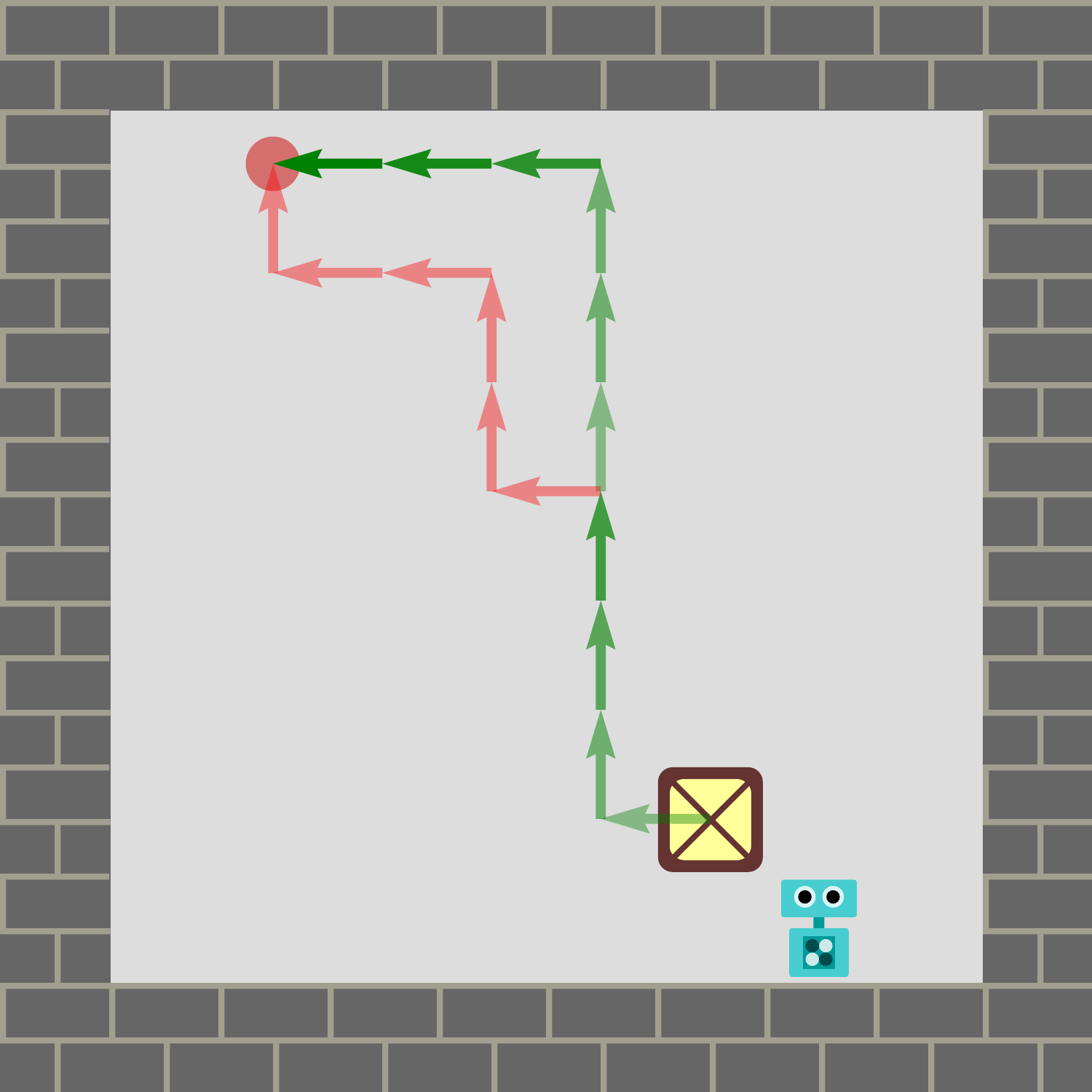}%
    \end{minipage}%
\end{figfullwidth}

We examined the Box-Directions probe, which displayed significantly higher causal influence than the Agent-Directions probe \cref{tab:causal-probe-results}. To better understand this behavior, we designed controlled scenarios. \Cref{fig:toy-levels-intervention} (left) shows a level where a single box is next to a target, and the agent faces two equally viable paths. Without intervention, the agent initially chooses the right, taking three steps before returning to the start and switching to the left. This behavior deviates from optimality, which requires committing to a single path.

When we applied Agent-Directions probe to enforce a path choice, the agent consistently followed the \emph{chosen trajectory} \cref{fig:toy-levels-intervention} (middle). However, in levels where multiple viable paths exists, the Agent-Directions probe's influence diminished, and the Box-Directions probe \emph{guided} the agent more effectively. \Cref{fig:toy-levels-intervention} (right) shows a near-empty level where the Box-Directions probe was used to guide the agent on the first four steps. After the intervention ceased. the agent computed a shorter, optimal path and deviated from the initial intervention.

This analysis the agent prioritizes box-directions when available and relies on the agent-direction cues as secondary guidance. These findings highlight the hierarchy of planning components encoded in the network.

\section{Case Studies}

\begin{figfullwidth}[h]{Case studies of three medium-validation levels demonstrating different behaviors after 6 thinking steps. Colors are as in \cref{fig:side-by-side-tinyworld} (right). Boxes and targets are paired in upper- and lower-case letters respectively, and the optimal solution places boxes alphabetically. Videos available at \href{https://drive.google.com/drive/folders/1qtxG5B_WGLHE2BusqkCBnpdSZCoQmYM6}{this https URL}. Levels solved faster incur fewer per-step penalties, yielding higher returns. Letters are for reference only and not intrinsic to Sokoban.\label{fig:case-studies}}%
\begin{minipage}{0.8\linewidth}%
\begin{subfigure}[t]{0.33\linewidth}%
\includegraphics[width=\linewidth]{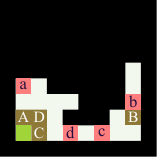}%
\caption{file 0, level 18\\unsolved $\overset{\text{think}}{\rightarrow}$ solved}%
\end{subfigure}%
\hspace{0.00499\linewidth}%
\begin{subfigure}[t]{0.33\linewidth}%
\includegraphics[width=\linewidth]{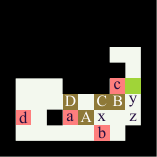}%
\caption{file 0, level 53\\$\overset{\text{think}}{\rightarrow}$ solved faster}%
\end{subfigure}%
\hspace{0.00499\linewidth}%
\begin{subfigure}[t]{0.33\linewidth}%
\includegraphics[width=\linewidth]{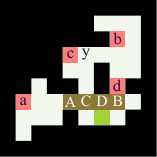}%
\caption{file 0, level 153\\$\overset{\text{think}}{\rightarrow}$ solved slower}%
\end{subfigure}%
\end{minipage}
\end{figfullwidth}

\paragraph{Case Study: Thinking makes some levels solvable \cref{fig:case-studies}(a).} In this scenario, the DRC fails to solve the level in the no-thinking condition. It initially pushes box $C$ one square to the right. While attempting to push $A$ to $a$, it blocks $B$ from reaching $b$, making the level unsolvable. In the thinking condition, the DRC correctly pushes $A$ to $a$ first, leaving room to push $B$ to $b$ and $C$ to ts target, solving the level.

\paragraph{Case Study: Thinking speeds up solving \cref{fig:case-studies}(b).} 

Without thinking, the DRC takes inefficient route, moving back and forth multiple times before starting to push boxes. For example, it moves to $y$, up to $c$, then down onto $z$, back up to $y$ and to $z$ before solving the puzzle. With thinking, the DRC immediately moves to box $A$, completing the solution with fewer extraneous steps. The resulting solution is faster, reducing the total step penalty.

\paragraph{Case Study: Thinking slows down \cref{fig:case-studies}(c).}  In this level, thinking causes the DRC to take a less efficient path. Without thinking, the DRC pushes box $C$ to $y$, followed by boxes $A$, then $B$ into place. On the way back down, the DRC pushes $C$ onto $c$ and finally $D$ onto $d$. In the thinking condition, the DRC starts by pushing B onto b then backtracks to push A, C, then D. This unnecessary backtracking leads to a slower overall solution.

\section{Probe Training}
\label{app:probetraining}
\paragraph{Probe architecture}
We train multiple linear probes on the hidden states $h$ and cell states $c$ activations of the network. Probes are trained either on activations from individual layers or by concatenating activations across all three layers. Activations were collected as the model played through hard levels. The training set are constructed by randomly sampling cached levels and including all timesteps except the first five. These initial steps excluded due to potential noise while the network formulated its strategy for solving the levels.

Probes are trained with logistic regression with L1 decay using the Scikit-Learn library. A grid search over
learning rates and L1 weight decay was conducted to identify the probe with the highest F1 score on the validation set.

For multi-class probe targets, each potential output class $l$ is treated as a separate data point. Specificially: A prediction was considered positive if the highest probe logit corresponded to class $l$, and negative otherwise. A data point was positive if its true label matched $l$. The confusion matrix and F1 score were then computed 
from the $n \cdot l$ data points.

In addition to Agent-Directions and Box-Directions probes, we trained the following:
\begin{itemize}
  \item \textbf{Next-Box probe}: Predicts $1$ on the square of the box that the agent will move next and $0$ for every other square.
  \item \textbf{Next-Target probe}: Predicts $1$ on the square of the target that the agent will put a box in next, and $0$ for every other square.
  \item \textbf{Next-Action probe}: Grid-wise binary probe for each of the four move action that predict $1$ for all squares in the grid if the next action matches the probe's action. We train grid-wise probe instead of a global probe so that the probe can be applied to arbitrary-sized inputs. The results for the next-action probes are available in \cref{tab:action_features}.
  \item \textbf{Pacing probe}: The global label is 1 if the agent is currently in a cycle, and 0 otherwise.
  \item \textbf{Value probe}: The global label is the numerical value that the critic head outputs.
\end{itemize}

\paragraph{Probe results.}
The pacing probe gets $F_{1}=31.0\%$, which is not much better than the constant 1 probe, which has
$F_{1}=12.8\%$. This suggests that the \drcthree{} does not represent whether it is in a cycle or not.

For the value function probe, we compute the fraction of variance explained $R^{2}$. If we train a global probe, $R^{2}$
is very high: $97.7\%$--$99.7\%$ depending on the layer. However, grid-wise probes obtain much worse but still passable
results: $41.0\%$--$79.2\%$ depending on layer. We visually checked whether the grid-wise probe reads off the values of
different plans, but could not find any such pattern.

Almost all the performance of the global probe is recovered by training on the mean-pooled inputs: $95.2\%$--$99.5\%$.
It is likely that the global and mean-pooled value probes are indirectly counting the number of squares the agent will
step on, which almost fully determines the value, and we know is possible due to the future-direction probes.

The possible box directions probe got approximately $\approx 30\%$ accuracy, and it was 

\begin{tabhalfwidth}[ht]{Weight and bias for transforming action probes to predictions\label{tab:weight-bias-direction-probes}}
\begin{tabular}{lrrr}
\toprule
 & Mean & Max & Positive proportion \\
\midrule
Weight & 1.2086 & -0.0582 & 0.2070 \\
\bottomrule
\end{tabular} \\[1ex]
\begin{tabular}{lrrrr}
\toprule
 & Up & Down & Left & Right \\
\midrule
Bias & 0.3337 & -0.0921 & -0.0632 & -0.0539 \\
\bottomrule
\end{tabular}
\end{tabhalfwidth}

\section{Looking for interpretable features with Sparse Autoencoders (SAEs)}\label{app:sae}

\begin{figfullwidth}[t]{Visualization of some interpretable features from the SAE of last layer. These features also appear monosemantically in the channels. The precision, recall, and F1 score for the features are reported in \cref{tab:sae-best-feature-results}.\label{fig:sae-feature-vis}}
\includegraphics[width=\linewidth]{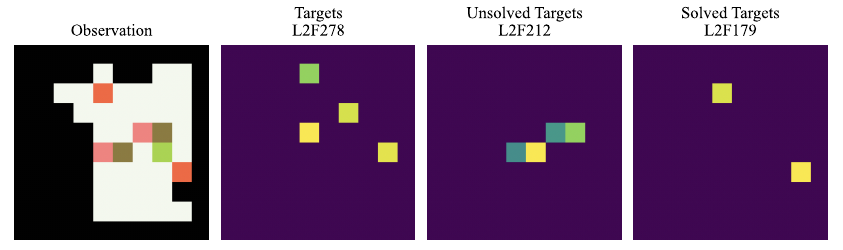}
\end{figfullwidth}

To search for monosemantic and interpretable features in the network, we train sparse-autoencoders (SAEs) \citep{huben2023sparse, bricken2023towards} on the individual squares in the $h$ hidden state of the network consisting of 32 neurons. Thus, we get a $10 \times 10$ visualization for each SAE feature as shown in Figure~\ref{fig:sae-feature-vis}. We use the top-k activation function \citep{gao2024scaling,tamkin2023codebookfeaturessparsediscrete}, which is state-of-the-art for training SAEs and directly enforces an $L_0$ sparcity constraint on activations. The SAE hyperparameter search space is detailed in Table~\ref{tab:sae-hyperparams}. We train separate SAEs for each layer with the specified hyperparameters, selecting those achieving greater than $90\%$ explained variance while retaining interpretable features based on manual visual inspection. We release these trained SAEs and probes in the huggingface repository with our trained DRC networks. \footnote{\url{https://huggingface.co/AlignmentResearch/learned-planner}}

Examples of interpretable features from an SAE trained on the last layer with $k=8$ are provided in \cref{tab:sae-feature-concepts}, with visualizations in \cref{fig:sae-feature-vis}. For ``Target'', ``Unsolved'', and ``Solved'' concepts (\cref{tab:sae-best-feature-results}), we occasionally observed an offset of 1 square (horizontal or vertical) from the ground truth. Due to a permanent one-square outer-edge wall, this offset never results in out-of-bounds errors. We evaluated these potential ``Offset'' variants for the Target, Unsolved, and Solved concepts.

All interpretable SAE features identified are already embedded in individual channels, with comparable or higher monosemanticity then the SAE features based on F1 scores.
\Cref{tab:action_features} compares precision, recall, and F1 scores for action features across
channels, SAE neurons, and linear probes trained against the ground truth predictions.
\Cref{tab:sae-best-feature-results} reports scores for other interpretable features. SAE action
features consistently underperform channels by an average F1 margin of $5.9\%$. Linear probes
trained across all hidden-state channels achieve similar F1 scores to individual channels indicating the
inherent monosemanticity of channels without further improvement through linear probe.

\begin{tabhalfwidth}[ht]{Hyperparameter search space for training SAE\label{tab:sae-hyperparams}}
\resizebox{\columnwidth}{!}{ 
\begin{tabular}{p{0.35\columnwidth}|p{0.6\columnwidth}} 
\toprule
\textsc{Hyperparameter} & \textsc{Search Space} \\
\midrule
$k$ & $\{4, 8, 12, 16\}$ \\
learning rate & $\{1e-5, 5e-5, 1e-4, 5e-4, 1e-3\}$ \\
expansion factor & $\{16, 32, 64\}$ \\
\bottomrule
\end{tabular}
}
\end{tabhalfwidth}

\begin{tabhalfwidth}[ht]{SAE Feature Concepts\label{tab:sae-feature-concepts}}
\begin{tabular}{l|l}
\toprule
\textsc{Concept} & \textsc{Description} \\
\midrule
Target & The 4 target squares (static) \\
Unsolved & Targets and boxes that aren't solved \\
Solved & Solved target squares with a box on them \\
Agent Up & The agent will move Up next step \\
Agent Down & The agent will move Down next step \\
Agent Left & The agent will move Left next step \\
Agent Right & The agent will move Right next step \\
\bottomrule
\end{tabular}
\end{tabhalfwidth}

\begin{tabhalfwidth}[ht]{Breakdown of levels by category at 6 thinking steps.\label{tab:behavior-breakdown}}
\begin{tabular}{lr}
\toprule
\textsc{Level categorization} & \textsc{Percentage} \\
\midrule
Solved, previously unsolved & 6.87 \\
Unsolved, previously solved & 2.23 \\
\midrule
Solved, with better returns & 18.98 \\
Solved, with the same returns & 50.16 \\
Solved, with worse returns & 5.26 \\
\midrule
Unsolved, with same or better returns & 15.14 \\
Unsolved, with worse returns & 1.36 \\
\bottomrule
\end{tabular}
\end{tabhalfwidth}

\begin{tabfullwidth}[htp]{Scores for SAE and Channel features\label{tab:sae-best-feature-results}}
\begin{tabular}{lr|lrrr|lrrr}
\toprule
\textsc{Concept} & \textsc{Offset $(dy,dx)$} &\multicolumn{4}{|c|}{\textsc{Channel}} &\multicolumn{4}{|c}{\textsc{SAE Feature}} \\
\midrule
& & \textsc{Number} & \textsc{Prec} & \textsc{Rec} & \textsc{F1}
& \textsc{Number} & \textsc{Prec} & \textsc{Rec} & \textsc{F1} \\
\midrule
Target & (1, 0) & 
L3C17 & 97.8 & 97.7 & 97.8 &
L3F278  & 97.8 & 98.1 & \textbf{98.0} \\
Unsolved targets and boxes & (0, 0) & 
L3C7 & 94.9 & 90.8 & \textbf{92.8} &
L3F212 & 95.3 & 86.6 & 90.7 \\
Solved targets & (0, 0) & 
-L3C7 & 91.6 & 94.6 & \textbf{93.0} &
L3F179 & 91.7 & 91.5 & 91.6 \\
\bottomrule
\end{tabular}
\end{tabfullwidth}

\begin{tabfullwidth}[hbtp]{Action features scores across channels, probes, and SAE features\label{tab:action_features}}
\begin{tabular}{l|lrrr|lrrr|rrr}
\toprule
\textsc{Feature}
&\multicolumn{4}{|c|}{\textsc{Channel}}
&\multicolumn{4}{|c}{\textsc{SAE Feature}}
&\multicolumn{3}{|c}{\textsc{Probe}}\\
\midrule
& \textsc{Number} & \textsc{Prec} & \textsc{Rec} & \textsc{F1}
& \textsc{Number} & \textsc{Prec} & \textsc{Rec} & \textsc{F1}
& \textsc{Prec} & \textsc{Rec} & \textsc{F1} \\
\midrule
Up& L3C29& 95.7& 88.1& \textbf{91.7}& L3F270& 93.9& 76.2& 84.1& 97.5& 86.5& \textbf{91.7}\\
Down& L3C8& 98.4& 80.8& 88.8& L3F187& 98.0& 79.1& 87.6& 97.6& 86.9& \textbf{91.9}\\
Left& L3C27& 85.5& 84.6& \textbf{85.1}& L3F244& 96.1& 63.2& 76.2& 83.5& 86.6& 85.0\\
Right& L3C3& 97.0& 86.9& 91.7& L3F385& 94.6& 78.5& 85.8& 97.6& 87.4& \textbf{92.2}\\
\bottomrule
\end{tabular}
\end{tabfullwidth}

\section{Additional quantitative behavior figures and tables}
\label{app:additional-figures}

\begin{figure*}[p]%
\begin{minipage}{0.45\textwidth}%
\includegraphics[width=\linewidth]{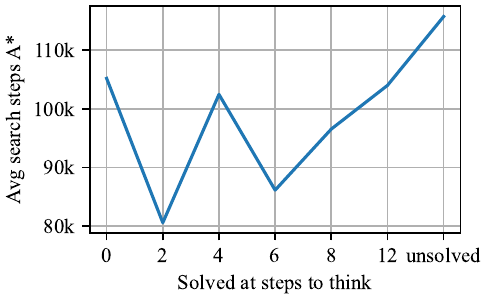}%
\caption{Number of thinking steps required to solve the level vs. number of nodes A* needs to expand to solve it. The weak correlation toward the end indicates different that the DRC and A* rely on different heuristics.\label{fig:thinking-steps-vs-nodes}}%
\end{minipage}%
\hspace{0.01999\textwidth}%
\begin{minipage}{0.49\textwidth}%
\includegraphics[width=\linewidth]{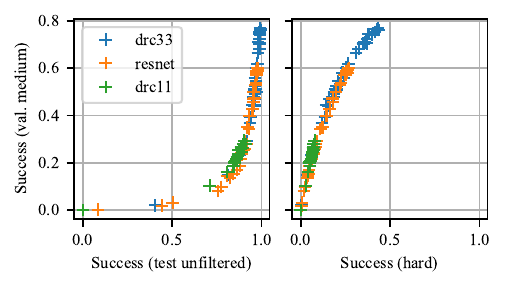}%
\caption{Success rate on datasets of varying difficulty for different architecture checkpoints. Performance trends suggest ResNets and DRCs with comparable results on easier sets also perform similarly on harder sets. $\drcone{}$ is a slight exception, performing consistently worse overall (see \cref{fig:test-valid-learning-curves}).\label{fig:relative-performance-on-hardness}}%
\end{minipage}
\end{figure*}

\Cref{fig:thinking-steps-vs-nodes} through \cref{fig:cycle-illustration} present quantitative behavior, analyzing various aspects of the network's planning behavior and evaluation metrics.


\begin{figfullwidth}[hbtp]{We replace $N$-length cycles with $N$ thinking steps to examine state consistncy across subsequent timesteps. \emph{(a)} Histogram of cycle lengths in the medium-validation set. \emph{(b, c)} After replacing a cycle with the same length in thinking steps, are all the states the same for the next $x$ steps?\label{fig:cycle-distribution}}
\begin{subfigure}{0.32\linewidth}
\includegraphics[width=\linewidth]{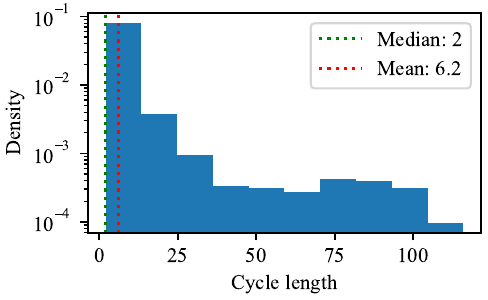}
\caption{Cycle length distribution}
\end{subfigure}
\hfill
\begin{subfigure}{0.32\linewidth}
\includegraphics[width=\linewidth]{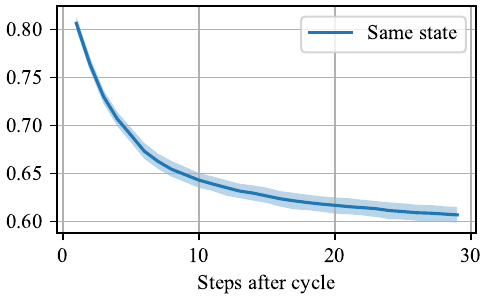}
\caption{with cycles from all levels}
\end{subfigure}
\hfill
\begin{subfigure}{0.32\linewidth}
\includegraphics[width=\linewidth]{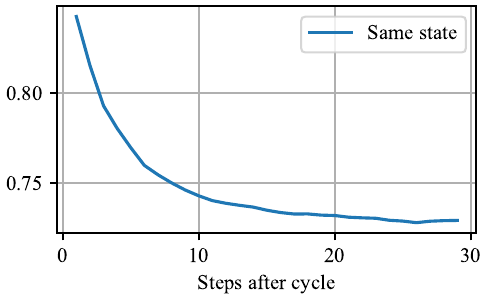}
\caption{with cycles from levels solved by agent}
\end{subfigure}
\end{figfullwidth}

\begin{figfullwidth}[hbtp]{\label{fig:cycle-illustration} Illustration of cycles and F1 scores}
\begin{subfigure}{0.4\linewidth}
\includegraphics[width=\linewidth]{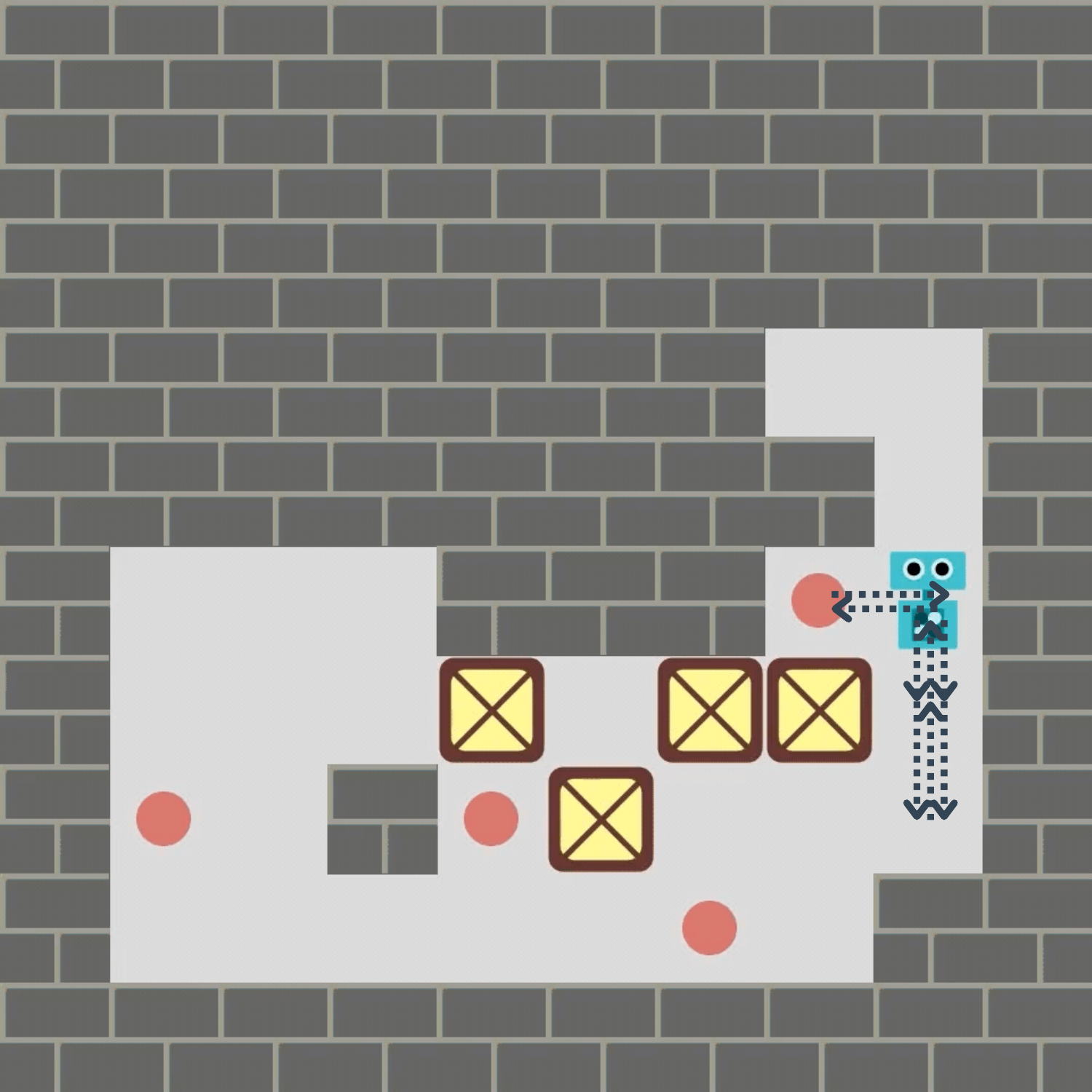}
\caption{Pacing behavior on file 0, level 53. On the given starting observation, the agent paces around 4 spaces in the first 9 steps and then goes on to solve the level. Video for the level is available at \ificlrfinal \href{https://drive.google.com/file/d/1M0Ebtc_1aAVgYV1Hki6FhLFMtDH57yRU/view}{this https url} \else this url (removed for double-blind review)\fi.}
\end{subfigure}
\hfill
\begin{subfigure}{0.59\linewidth}
\includegraphics[width=\linewidth]{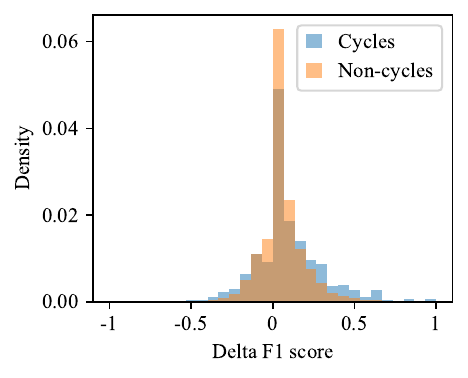}
\caption{Change in per-step F1 score of Box-Directions probe for moves in cycles and outside cycles on medium-difficulty validation levels. The non-cycle moves were recorded from the same distribution of timesteps where cycles occur but from levels without a cycle at those steps. Mean per-step change in F1 for cycle and non-cycle steps are $1.40\% \pm 0.06\%$ and $0.84\% \pm 0.04\%$ respectively.}
\end{subfigure}
\end{figfullwidth}

\section{Additional related work}\label{app:more-related-work}

\paragraph{Ethical treatment of AIs.} The question whether AIs deserve moral consideration has been widely debated. \Citet{schwitzgebel2015ai} argue that highly human-like AIs merit rights, while \Citet{tomasik2015suffering} suggest that most AIs deserve some degree of consideration, akin to biological organisms \citep{singer2004animal}. The concept of \emph{ethical} treatement  for reinforcement learners has been explored by \citet{daswani15happiness}, who propose that pleasure and pain in these systems may correspond to temporal difference (TD) error rather than absolute returns. If neural networks have internal objectives distinct from their critic head \citep{hubinger2019risks,di2022goal}, identifying these objectives could provide higher-assurance methods for assessing AI goals compared to direct querying \citep{perez23_towar_evaluat_ai_system_moral}.

\paragraph{Chain-of-thought faithfulness.} The faithfulness of chain of thought reasoning in large language models (LLMs). Studies such as \citep{lanham2023measuring,pfau24_lets_think_dot_by_dot} investigate whether LLMs rely on 
plain English for long-term reasoning, which could allow unintended consequences to be easily identified and mitigated \citet{scheurer23_large_languag_model_can_strat}.

\paragraph{Fully reverse engineering small networks.} Recent efforts have successfully reverse-engineered small neural networks performing algorithmic tasks
\citep{nanda2023progressmeasuresgrokkingmechanistic,chughtai23_toy_model_univer,zhong2023pizza,quirke2023understanding}.

\paragraph{Systematic Generalization.} Previous work identified conditions under which neural networks generalize, such as diverse datapoints and egocentric environments \citep{humanlikesysgen,Hill2020Environmental,mutti2022invariance}. Similar interpretability can be extended across these neural networks to uncover shared planning mechanisms and the conditions in which they emerge.

\section{Sussman's anomaly}

\begin{figure}
    \hspace{0.7in}
    \includegraphics[width=0.8\linewidth]{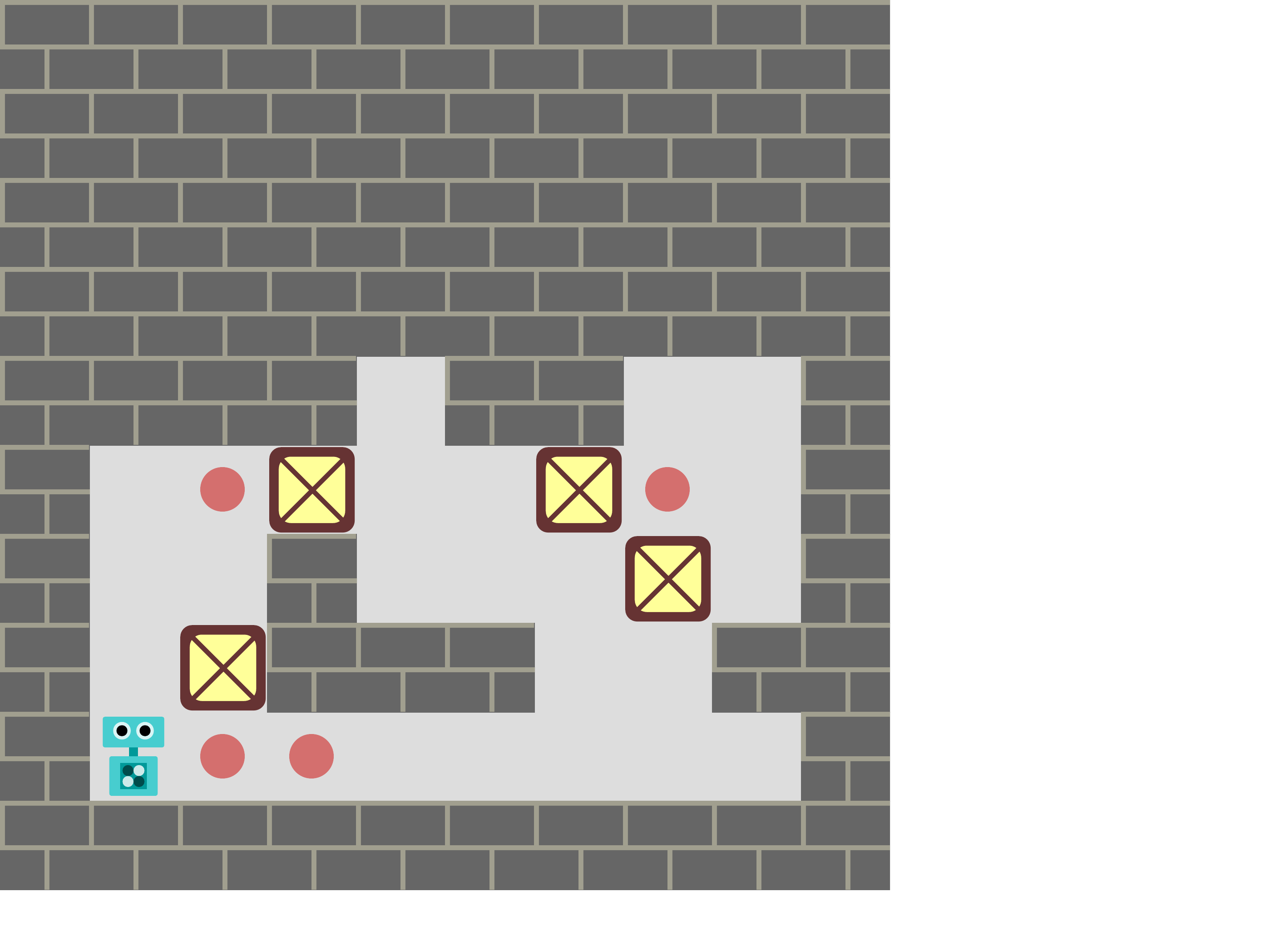}
    \caption{Level 466 from file 233 of train-medium in which the \drcthree{} network tries to put both the boxes on the right to the target on the right by repeatedly putting one box on target, only to remove it and put the other one on the same target.}
    \label{fig:sussman}
\end{figure}

Although the \drcthree{} shows exceptional long-term planning capabilities as demonstrated in the paper, it can have some trivial failure modes. \Cref{fig:sussman} shows a level in which the agent indefinitely tries to put the two boxes on the right onto the same target that is on the right side of the level. This failure mode is similar to the Sussman's anomaly demonstrated by \citet{sussman_1973} illustrating the weakness of non-interleaved planning algorithms. We observe that such cases happen rarely, with the network being able to resolve the interleaved dependencies between boxes in most cases.

\section{Planning vs. predicting box-directions}

Most of our analysis and experimentation in the paper relies on the box-directions probe, which predicts the sequence of all future box movement directions from squares in the grid given network's activations. While predicting the box-directions is very close to planning, it misses a few details required for a complete plan. For example, for a complete plan of actions, the agent also has to also know the order in which it wants to move the boxes in and the plan a path from the agent's current position to the next box to push. However, we do train probes for agent-directions and next-box to move and show they have high predictive accuracy. We focus primarily on boxes-directions because it is the most crucial component of a plan in the game, which is also reflected in the fact that the box-directions probe is the most causal probe.


\end{document}